\documentclass[10pt,twocolumn,letterpaper,table]{article}
\usepackage{arxiv}
\usepackage{appendix}
\usepackage{booktabs}
\usepackage{times}
\usepackage{latexsym}
\usepackage{amsmath}
\usepackage{xcolor}
\usepackage{graphicx}
\usepackage{enumitem}
\usepackage{multirow}
\usepackage{subcaption}
\usepackage{mathtools}
\usepackage{algorithm}
\usepackage{algpseudocode}
\usepackage{bm}
\usepackage{adjustbox}
\usepackage{amsmath}
\usepackage{tabularray}

\newcommand{\para}[1]{\noindent\textbf{#1.}}
\newcommand{\paraq}[1]{\noindent\textbf{#1?}}
\definecolor{lgreen}{RGB}{236, 255, 201}
\DeclareMathOperator*{\argmax}{arg\,max}

\usepackage{hyperref}
\usepackage[capitalize]{cleveref}
\crefname{section}{Sec.}{Secs.}
\Crefname{section}{Section}{Sections}
\Crefname{table}{Table}{Tables}
\crefname{table}{Tab.}{Tabs.}

\hypersetup{
    colorlinks=true,
    linkcolor=magenta,
    urlcolor=cyan,
    }

\begin{document}
\title{SSE: Multimodal Semantic Data Selection and Enrichment \\
for Industrial-scale Data Assimilation}

\author{
Maying Shen\thanks{Equal contribution.}, \ Nadine Chang\footnotemark[1], \  Sifei Liu, \ Jose M. Alvarez \\
NVIDIA \\
{\tt\small\{mshen, nadinec, sifeil, josea\}@nvidia.com}
}

\maketitle

\begin{figure*}[!th]
    \centering
    \begin{subfigure}[t]{0.33\textwidth}
        \centering
        \includegraphics[width=\linewidth]{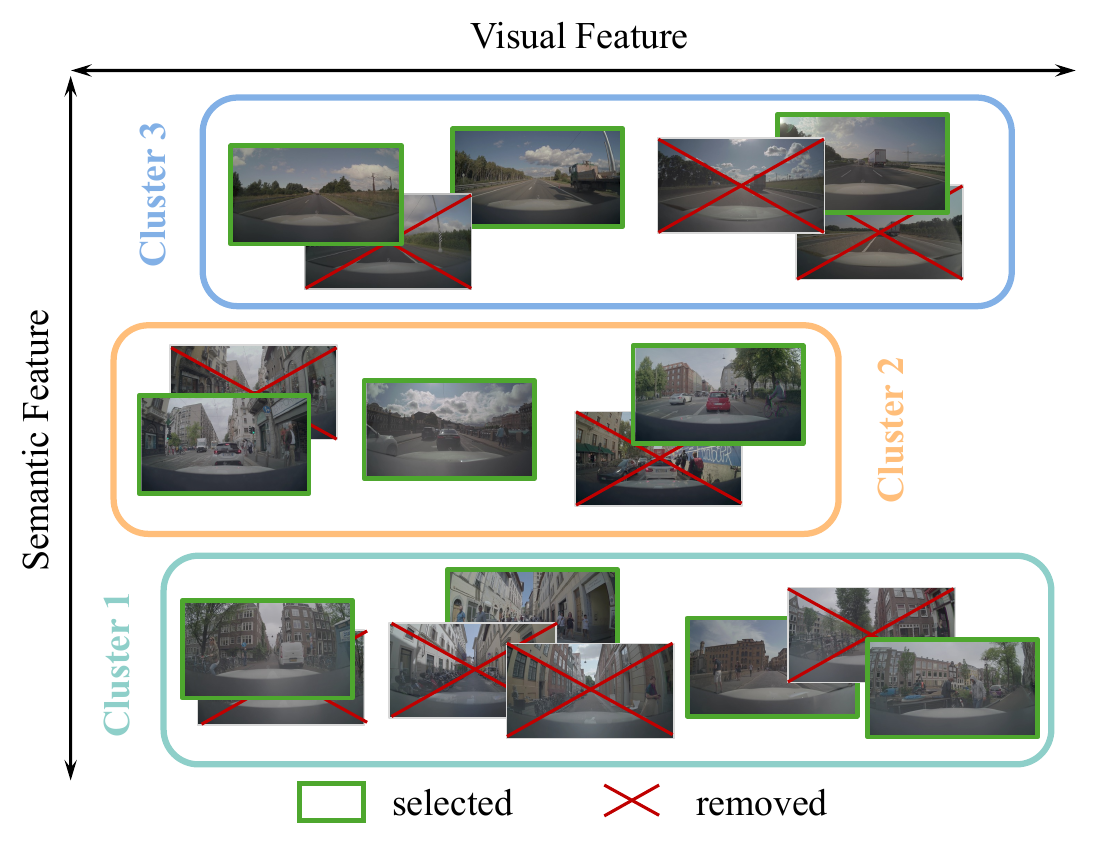}
        {(a) Semantic data selection}
    \end{subfigure}%
    ~
    \begin{subfigure}[t]{0.28\textwidth}
        \centering
        \includegraphics[width=\linewidth]{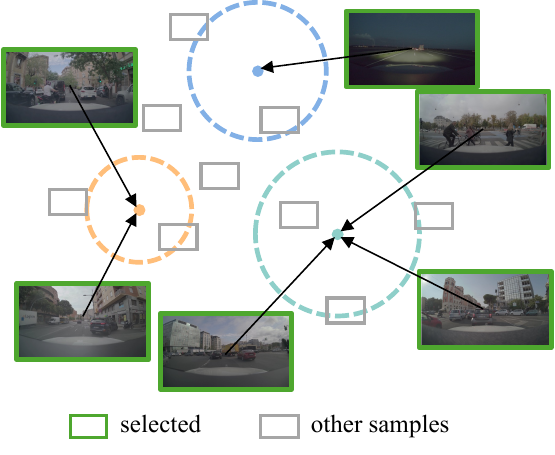}
        {(b) Semantic data enrichment}
    \end{subfigure}%
    ~
    \begin{subfigure}[t]{0.35\textwidth}
        \centering
        \includegraphics[width=\linewidth]{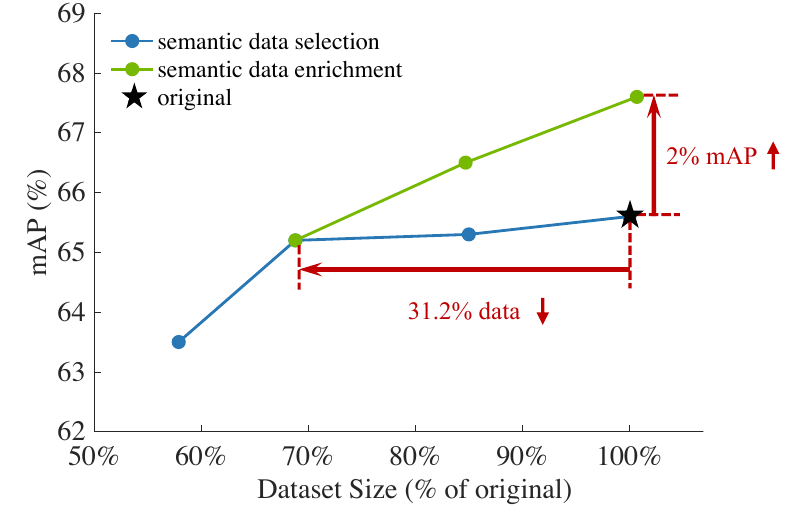}
        {(c) 3D object detection results with SSE}
    \end{subfigure}
    % \vspace{-2mm}
    \caption{
    We introduce our semantic data selection and enrichment framework (SSE) for autonomous vehicles. The framework generates semantic captions for each data point using a foundation model, capturing semantics including scene understanding (e.g., ``crowded urban intersection") and crucial object interactions (e.g., ``person about to cross in front of car"). (a) To create a compact dataset, we select the most semantically important portions of a curated and labeled dataset, removing visually similar scenes. (b) To enrich the dataset, we identify new important data points, which are semantically distant from our labeled dataset, from a growing unlabeled data pool. (c) With this approach, we maintain downstream 3D object detection performance using only 70\% of the labeled dataset, and we can enhance model performance without increasing the original training dataset size by enriching the selected dataset. }
    \label{fig:teaser}
\end{figure*}

\begin{abstract}

In recent years, the data collected for artificial intelligence has grown to an unmanageable amount. Particularly within industrial applications, such as autonomous vehicles, model training computation budgets are being exceeded while model performance is saturating -- and yet more data continues to pour in. To navigate the flood of data, we propose a framework to select the most semantically diverse and important dataset portion. Then, we further semantically enrich it by discovering meaningful new data from a massive unlabeled data pool. Importantly, we can provide explainability by leveraging foundation models to generate semantics for every data point. We quantitatively show that our Semantic Selection and Enrichment framework (SSE) can a) successfully maintain model performance with a smaller training dataset and b) improve model performance by enriching the smaller dataset without exceeding the original dataset size. Consequently, we demonstrate that semantic diversity is imperative for optimal data selection and model performance. 

\end{abstract}

\section{Introduction}

The advent of successful artificial intelligence models have led to an exponential data growth in recent years. As we continue to collect, label, and train on ever expanding data, larger and more impressive models are being created, which recently culminated in foundation models. However, this insatiable data growth is now leading to two new challenges. 1) Our labeled datasets are so large that model performances have saturated. 2) The continuous stream of unlabeled data needs to be properly filtered in order to discover the most valuable data, like a needle in a haystack. Both of these challenges are especially pronounced within the industrial autonomous vehicle (AV) field, as data sizes are much larger due to big fleets that collect large-scale multi-sensor, temporal and 3D data. In this paper, we address these two particular challenges in a task agnostical manner and showcase results in the context of AV.

The first challenge when working with an excessive data amount is that training on all labeled data is computationally unaffordable. Under a reasonable compute cost, studies have shown that more data does not equate to better performance. Instead, we need less data that is higher quality for optimal model performance~\cite{goyal2024scaling}. This discovery has spurred an emerging new branch of literature: data selection/pruning. Hence, the second challenge consists of finding valuable data in a pool of unlabeled and unorganized data. A field of academic study that is related to this challenge is known as active learning, which primarily focuses on improving early stages of training. However, data pools in active learning are much smaller than the growing avalanche of data we see today. Therefore, we define this second challenge as data enrichment. Additionally, both data selection and enrichment would benefit from additional explainability, as knowing why data points were selected helps to verify their inherent value.

\para{Data Selection}
The emerging field of data selection aims to reduce an existing labeled dataset into a compact dataset. Recent studies achieve this by targeting high-quality and pruning low-quality data, mainly in classification and detection datasets. They identify high-quality datasets as either having a traditional object-balanced distribution, visual diversity, or semantic diversity. For traditional object distribution balancing, works refer to long-tail approaches that upsample rarer objects as overall object balancing strategy. Existing released datasets, such as NuScenes~\cite{caesar2020nuscenes}, incorporate a variety of these statistical dataset balancing approaches to produce a reasonably sized clean dataset. For visual diversity, other works maximize visual embedding coverage space~\cite{sener2017active}.

Since balanced dataset statistics do not guarantee a meaningful content diversity, other works pivot to creating semantically diverse datasets. Within the context of AV, semantic diversity is defined as the overall scene diversity across one driving scenario, which may include the following: A description of the overall background scene, dynamic objects, car behaviors, weather or presence of hazardous objects. Recent works leverage the world knowledge encapsulated by foundation models like CLIP~\cite{radford2021CLIP} to encode images into stronger embedding spaces with more semantic information than previous vision embeddings, and then remove images with similar embeddings. Thus, they mostly focus on visual duplicate removal as opposed to semantically relevant data-point selection. 

\para{Data Enrichment}
Within academia, parts of data enrichment fall under active learning. Similar to data pruning, active learning has been studied mostly for image classification. While some works consider 2D object detection, they often focus on early stages of the training with a small amount of data~\cite{choi2021active,kao2019localization}. Broadly, active learning uses specialized models, uncertainty estimation, or query by committee. Specialized models are trained to identify targeted objects (e.g. a ``traffic light’’ detector)~\cite{shu2023knowledge,bertoni2017data}, but this methodology does not scale and cannot respond to new targets. Uncertainty estimation utilizes the downstream model to measure data sample uncertainty to determine which new samples to select~\cite{lakshminarayanan2017simple,liu2020simple,valdenegro1910deep}. As AV includes several constantly evolving downstream models and multi-sensor data, relying on a single model and specialized, model-specific data for uncertainty estimation is impractical. Query by committee uses majority vote by model ensembles, which requires even more model development and computational costs~\cite{roy2018deep}. Unlike in data pruning, these works also do not focus on data semantics. 

\para{Common Issues}
Existing works have two main issues: First, they lack explainability, which is a shared issue across data selection and enrichment works. Describing the contents of selected data points is critical to safety and helps us understand why some data are more important than others. Second, existing works disjointly study data selection and enrichment. But realistically, both have to be considered simultaneously. We emphasize that within industry, data selection cannot be performed across the labeled and unlabeled data pools because the unlabeled pool increases and changes continuously. Thus, data selection is restricted to human-labeled datasets, which can be separately enriched with new data from the unlabeled data pool. To address the lack of explainability and disjoint processes, as shown in~\cref{fig:teaser}, we propose a joint semantic data selection and enrichment framework (SSE) that leverages multimodal language models (MLLMs) to generate explainable semantics in natural language.

\para{Our Framework (SSE)} Given an existing hand-curated and labeled dataset, we first focus on selecting the most relevant data without relying on labels. We define semantically diverse data points as the most relevant ones. Semantics are achieved by prompting MLLMs to describe each data sample in detail -- from producing overall scene descriptions (intersection in city) to pointing out notable hazards (bicyclists potentially crossing). These captions are then embedded and clustered. As seen in~\cref{fig:teaser}, we aim to select the most semantically diverse samples. Therefore, we organize them into core semantic clusters and subsequently remove visually similar samples within each cluster. After data selection, we enrich our dataset with additional semantically diverse scenes from a new \textit{unlabeled} data source(~\cref{fig:teaser}). Importantly, our methodology does not require any human annotated data labels. 

\para{Contributions} With our joint semantic data selection and data enrichment framework, we address the data challenges unique to industrial scale AV: 
\begin{enumerate}[leftmargin=*,topsep=0.15\baselineskip]
\setlength{\itemsep}{.15\baselineskip}%
\setlength{\parskip}{0pt}%measure (
\setlength{\topsep}{0pt}
\setlength{\labelindent}{0pt}%
    \item We leverage MLLMs to inject explainability into our process and remove reliance on downstream models. 
    \item We propose a joint framework to incorporate both data selection and enrichment(SSE) to address the two challenges arising from exponential data growth. 
    \item In addition to improving semantic selection and enrichment, we also provide a better semantic retrieval that uses our semantic embeddings. 
    \item We conduct thorough evaluations and analysis on large-scale industrial datasets and demonstrate downstream model performance improvements over baseline methodologies.
	\item We experimentally show that a semantically tuned dataset with fewer but more semantically relevant class objects is more effective for model performance than a hand-curated and statistically tuned dataset with strictly more class objects. 
\end{enumerate}

\section{Method}

A high-level overview of our semantic selection and enrichment frame is shown in~\cref{fig:teaser}. An additional pseudocode is provided in~\cref{pseudo}.

\algnewcommand\algorithmicforeach{\textbf{for each}}
\algdef{S}[FOR]{ForEach}[1]{\algorithmicforeach\ #1\ \algorithmicdo}

\newcommand{\myindent}[1]{
\newline\makebox[#1cm]{}
}

\begin{algorithm}[th]
\caption{Data Selection and Enrichment} 
\label{pseudo}
\begin{algorithmic}[1]
\Require labeled dataset, $D$; unlabeled data pool, $P$
\Require prompt, $u$; pruning threshold, $\epsilon$
% the words associated with $c$, $L_c$; the training index $D_{tr}$
\Ensure $D_s$, reduced dataset after data selection and $D_e$, enriched dataset after data enrichment

%Convert images to captions with user prompt
\State Let $C^D = \{\mathrm{MLLM}(u, i): i \in D\}$ \label{op1} 

%Convert captions to text embedding
\State Let $T^D = \{\mathrm{SenTrans}(c_i): c_i \in C^D\}$ \label{op2}

\State Let clusters $M = \mathrm{kMeans}(T^D)$ \label{ds_start}

%Convert captions to text embedding
\State Let $V^D = \{\mathrm{CLIP}(i): i \in D\}$

%Data Selection
\ForEach{$m \in M$}
    \ForEach{$i \in m$}
        \If {$1-\mathrm{cos}(v_i, v_j) < \epsilon$,  \myindent{.9} for any other sample $j \in m, i\neq j$}
            \State Remove $j$ from $m$
        \EndIf
    \EndFor
\EndFor

\State $D_s = \{i: \exists i\in m \} $ for $m \in M$ \Comment{Selection} \label{ds_end}

\item[]
%Data enrichment
\State Let $C^P = \{\mathrm{MLLM}(u, i): i \in P\}$ \label{de_op1} 
\State Let $T^P = \{\mathrm{SenTrans}(c_i): c_i \in C^P\}$ \label{de_op2}
% \State Define $o$ as image of embedding closet to centroid: $m \in M$
\State Define $O = \{{\mathrm{ImageClosestCentroid}(m): m \in M }\}$ \label{de_start}

\State $D_e = D_s$
\While{expanding $D_e$}  \Comment{Enrichment}
    \State $D_e$ = $D_e$ + $\argmax_{i \in P} (1-\mathrm{cos}(t_i, t_o)): o \in O$
\EndWhile \label{de_end}

\end{algorithmic}
\end{algorithm}

\para{Capturing Semantics within Scenes}
We define the semantics within a single scene as answers to the following description requests: a general scene description, a general description of what is happening, the important objects to consider while driving or the dynamic objects. All descriptions require world knowledge and high levels of scene understanding. Thus, we ask a number of questions regarding a single image to a pre-trained MLLMs and receive a paragraph of desired answers. By encoding this caption paragraph into a text embedding space via a sentence transformer, we capture high-order semantics for a single scene. This process is illustrated for the labeled dataset, \textit{D}, and unlabeled data pool, \textit{P},  in~\cref{op1,op2}, \cref{de_op1,de_op2}.

\para{Semantic Data Selection}
Semantic embeddings enable identifying the various unique semantics within the dataset. As our goal is semantic diversity within a dataset, we perform \textit{k}-means clustering on the semantic embeddings to group semantically similar scenes. To select a core group of semantic scenes, we remove visually similar scenes within each semantic clusters. To calculate visual similarity, we leverage another foundation model (CLIP) and its vision encoder to obtain visual embeddings for each sample. Within each semantic cluster, we calculate the pairwise cosine similarity between the visual embeddings for all samples and greedily remove those whose score exceeds a pruning threshold $\epsilon$. The final dataset, $D_s$, contains the remaining samples within each semantic cluster. More detailed pseudocode is shown from ~\cref{ds_start} to ~\cref{ds_end}.
% \para{Removing Visually Similar Scenes}

\para{Semantic Data Enrichment}
\label{sec:method_de}
After semantic data selection, we can expand the dataset by adding new semantically meaningful data from an unlabeled data pool, \textit{P}. Since we have a set of semantic clusters, we utilize the semantic embedding closest to the centroid of each cluster as a semantic anchor point. Using these semantic anchors, we identify the data points within \textit{P} that are most semantically removed from our current semantics. We identify these points by calculating the cosine similarity between semantic embeddings. Then, we incrementally add data until the desired amount is reached. Finally, we annotate these data points analogous to the labeled dataset. The semantic enrichment is detailed from~\cref{de_start} to ~\cref{de_end}.

\section{Experiments}

\begin{table}[t!]
    \centering
    \caption{Performance on downstream 3D detection with different \underline{\textnormal{Data Selection}} strategies. SSE achieves $30\%$ data reduction while maintaining original mAP.}
    % \vspace{-0.3cm}
    \begin{minipage}{\columnwidth}
    \resizebox{.99\columnwidth}{!}{
    \begin{tabular}{ccccc}
         \toprule
        Dataset Size & \multirow{2}{*}{Method} & \multirow{2}{*}{mAP} & \multirow{2}{*}{$\Delta$\textunderscore original} & \multirow{2}{*}{$\Delta$\textunderscore random} \\
        ($\%$ of original) & & & & \\
        \midrule
        $100\%$ & Original Dataset & $65.6$ & - & - \\
        \midrule
        \multirow{4}{*}{$80\%$} & Random & $63.4$ & -$2.2$ & - \\
         & Long-tail~\cite{gupta2019lvis} & $64.2$ & -$1.4$ & +$0.8$ \\
         & CLIP visual~\cite{radford2021CLIP} & $64.9$ & -$0.7$ & +$1.5$ \\
         & \textbf{SSE (Ours)} & $\mathbf{65.1}$ & -$\mathbf{0.5}$ & +$\mathbf{1.7}$ \\
        \midrule
        \multirow{4}{*}{$70\%$} & Random & $62.1$ & -$3.5$ & - \\
         & Long-tail~\cite{gupta2019lvis} & $62.3$ & -$3.3$ & +$0.2$ \\
         & CLIP visual~\cite{radford2021CLIP} & $64.4$ & -$1.2$ & +$2.3$ \\
         \rowcolor{lgreen}
         & \textbf{SSE (Ours)} & $\mathbf{65.2}$ & -$\mathbf{0.4}$ & +$\mathbf{3.1}$ \\
         \midrule
         \multirow{4}{*}{$60\%$}&  Random & $61.1$ & -$4.5$ & - \\
         & Long-tail~\cite{gupta2019lvis} & $60.8$ & -$4.8$ & -$0.3$ \\
         & CLIP visual~\cite{radford2021CLIP} & $62.9$ & -$2.7$ & +$1.8$ \\
         & \textbf{SSE (Ours)} & $\mathbf{63.5}$ & -$\mathbf{2.1}$ & +$\mathbf{2.4}$ \\
        \bottomrule
    \end{tabular}
    }
    \end{minipage}
    \label{tab:main_selection}
    \vspace{-0.3cm}
\end{table}

\begin{figure*}[t!]
\centering
\includegraphics[width=\linewidth]{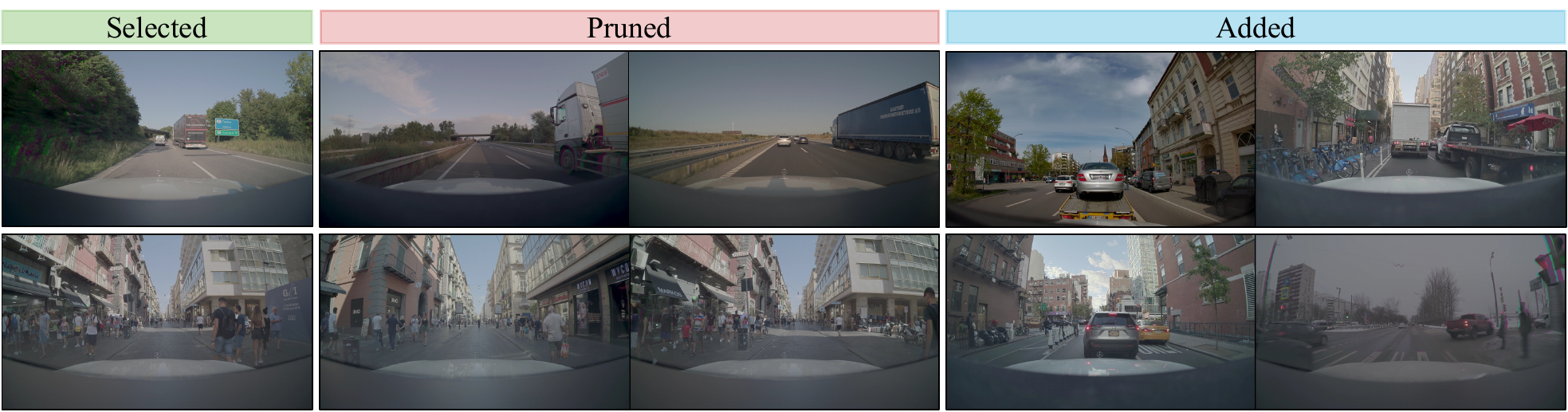}
\caption{ Examples of semantic selection and enrichment. The ``Pruned'' samples are visually and semantically similar to the "Selected" samples, not only a visual duplication. The ``Added'' samples add different semantics to existing data.}

\label{fig:main_vis}
\end{figure*}

\subsection{Implementation Details}
\para{Data}
We utilize an internal large-scale experimental research dataset \textit{D} with $415,544$ unique scenes. Each scene is $1$ driving timepoint recorded with $8$ cameras ($4$ standard and $4$ fisheye) and belongs to a full video sequence, which we define as a session. \textit{D} contains $2750$ unique sessions. The images are reduced to $480\times960$ resolution for training and evaluation. For data selection, we select unique scenes from the dataset \textit{D}. For data enrichment, we select unique scenes from an unlabeled data pool \textit{P} containing $662,325$ unique scenes from $5109$ unique sessions. 

\para{Evaluation Model}
We evaluate our semantic data selection and enrichment pipeline on the downstream Multi-Camera 3D Object detection model~\cite{pham2023nvautonet}. For the evaluations, we train the model for $20$ epochs with data from all $8$ cameras and an individual batchsize $8$ on $32$ GPUs in total. The learning rate is scheduled according to the one-cycle learning rate policy~\cite{smith2019super} with a maximum value at $5e^{-3}$. All final metrics are expressed as mean average precision (mAP), and per-class APs are reported in Appendix.

\para{Data Selection and Enrichment}
We use LLaVA1.6-34B-4bit ~\cite{liu2023improvedllava} to generate semantics, because it can handle high-resolution images. Unless specified, our semantic selection and enrichment utilizes MPNet-base-v2~\cite{song2020mpnet} for semantic caption encoding and $k=300$ for clustering, the front camera image as input to MLLM, and the specialized AV prompt in~\cref{fig:prompt_sensitivity} for generating image descriptions.

\subsection{Main Results: Semantic Selection and Enrichment Performance}
\para{Data Selection}
\label{sec:ds}
Our pipeline begins with data selection, where we aim to select a subset of semantically relevant data from an established, human-labeled internal dataset, \textit{D}. We compare against three main baselines, as seen in~\cref{tab:main_selection}. As an effective baseline, the random baseline randomly selects a portion of dataset to retain without any clustering. Next, following standard practice in dataset collection, we select a portion with the most balanced object distribution. We use Repeat Factor Sampling (RFS)~\cite{gupta2019lvis}, a strong standard baseline in long-tail detection for calculating a data sample's importance based on object count. Finally, our methodology leverages semantic scene understanding by converting generated image captions into text embeddings. To evaluate its effectiveness, we compare against the traditional approach of encoding images into visual embeddings. For this, we adopt CLIP~\cite{radford2021CLIP}, a powerful and widely adopted vision encoder, as a robust visual baseline. Once images are clustered based on both Visual and Semantic embeddings, data selection is performed by greedily pruning the most visually similar images within each cluster. 

Industry applications have to address the tradeoff between dataset retention percentage and performance. Usually, they select the method with the most aggressive dataset retention and negligible performance change. Practically, this decision translates to the cheapest and most efficient model training for maximum performance effect. In ~\cref{tab:main_selection}, we observe that our semantic data selection selects the smallest required amount of data (70\%) to achieve comparable performance to using the entire dataset. The difference of $0.4$ mAP is negligible. In comparison, the performance of other baselines drops noticeably when more data is pruned from the dataset. Note, that while object balancing seems effective for general long-tail detection settings, this strategy does not perform well when selecting relevant complex AV scenes.  

\begin{table}[th]
    \centering
    \caption{Performance on downstream 3D detection with different \underline{\textnormal{Data Enrichment}} methods. SSE improves 2 mAP when expanding the dataset to the original size. Note long-tail is impossible due to the lack of labels in data pool.}
    % \vspace{-3mm}
    \begin{minipage}{\columnwidth}
    \resizebox{.99\columnwidth}{!}{
    \begin{tabular}{ccccc}
         \toprule
        Dataset Size & \multirow{2}{*}{Method} & \multirow{2}{*}{mAP} & \multirow{2}{*}{$\Delta$\textunderscore original} & \multirow{2}{*}{$\Delta$\textunderscore random} \\
        ($\%$ of original) & & & & \\
        \midrule
        $100\%$ & Original Dataset & $65.6$ & - & - \\
        \midrule
        \multirow{3}{*}{$85\%$} & Random & $64.5$ & -$1.1$ & - \\
         & CLIP visual~\cite{radford2021CLIP} & $65.6$ & $0.0$ & +$1.1$ \\
         & \textbf{SSE (Ours)} & $\mathbf{66.5}$ & +$\mathbf{0.9}$ & +$\mathbf{2.0}$ \\
         \midrule
         \multirow{3}{*}{$100\%$} & Random & $65.2$ & -$0.4$ & - \\
         & CLIP visual~\cite{radford2021CLIP} & $65.6$ & +$0.4$ & +$0.8$ \\
         \rowcolor{lgreen}
         & \textbf{SSE (Ours)} & $\mathbf{67.6}$ & +$\mathbf{2.0}$ & +$\mathbf{2.4}$ \\
        \bottomrule
    \end{tabular}
    }
    \end{minipage}
    \label{tab:main_enrich}
    \vspace{-2mm}
\end{table}

\paraq{Do We Only Prune Identical Scenes}
In~\cref{fig:main_vis}, we visually demonstrate that the pruned data is similar to the selected sample, but importantly \textit{not} a visual duplication. For example, the first row shows that our pruning removed freeway scenes with trucks, which are similar to the selected data point. 

\begin{figure}[t]
    \centering
    \includegraphics[width=\linewidth]{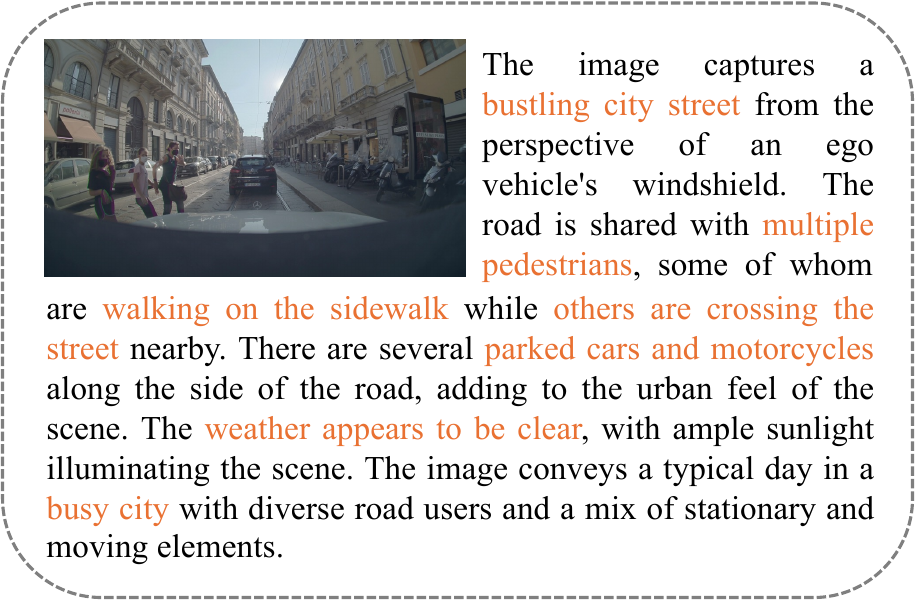}
    \caption{Semantic description with MLLMs. The highlighted phrases capture the relevant semantics.}
    \vspace{-4mm}
    \label{fig:interp}
\end{figure}

\begin{figure*}[t!]
    \centering
    \includegraphics[width=\linewidth]{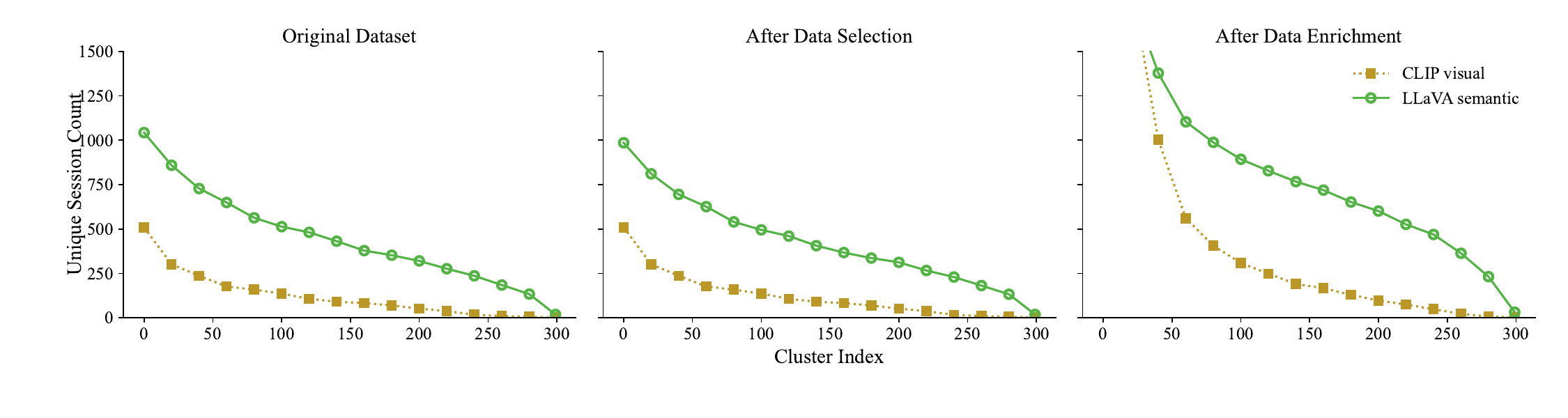}
    \caption{Number of unique driving video sessions in each cluster formed with different embeddings. Compared to clusters generated from visual embeddings, semantic clusters capture more semantically similar yet visually diverse scenes across sessions.}
    \vspace{-0.2cm}
    \label{fig:unique_sessions}
\end{figure*}

\para{Data Enrichment}
Continuously collected unlabelled data is a potentially useful addition to an existing dataset.
Here, we demonstrate the effect of applying our methodology on data enrichment for a carefully pruned and labeled dataset, \textit{$D_s$}. 
We apply data enrichment on top of all previous methodologies and their respective $70\%$ retained datasets. As we do not have a labeled pool of data to work with, we cannot report a long-tail enrichment baseline. For the remaining enrichment results, the methodology used for data selection is also used for data enrichment. For example, in the CLIP visual baseline, we encode the unlabeled data with the CLIP vision encoder and apply our semantic selection method, as detailed in~\cref{sec:method_de}. Once data is selected from the unlabeled data pool \textit{P}, we label the data following the same labeling guidelines used in our original dataset and expand \textit{D} for training.

We observe in~\cref{tab:main_enrich} that by growing a dataset with semantic embeddings only up to $85\%$ of the original dataset size, we are able to outperform the model trained on the original full dataset ($0.9$ mAP). By semantically expanding the dataset to the original dataset size, we are able to even push the improvement to 2 mAP.

\para{Explainability Enabled by Semantics}
Because our semantics are from generated captions, we can analyze why a data point is semantically important. The highlighted phrases in the generated captions shown in~\cref{fig:interp} succinctly capture the semantics within the given image. As a result of the generated caption for this sample, we are able to verify and understand that its semantic embedding refers to ``bustling city street'', ``multiple pedstrains...walking on the sidewalk'' and ``several parked cars and motorcycles''. Thus, we can inject explainability into every aspect of SSE.

% \vspace{-0.2cm}
\subsection{Main Results: Semantic Importance}
\paraq{How Semantically Diverse Are the Scenes in Enrichment}
In~\cref{fig:main_vis}, we see that newly added data is visually and semantically different from the reference image in enrichment process. In the upper row, the new data points are non-highway scenes where the driving space is more restricted.

By leveraging the semantic embeddings, our methodology demonstrates two impactful results: First, through semantic selection, we can drastically reduce the size of a carefully human-curated and annotated dataset. Second, through semantic data enrichment, we obtain an effective measure to further grow this pruned dataset for further performance improvements. 

\begin{figure*}[ht]
    \centering
    \includegraphics[width=0.19\linewidth]{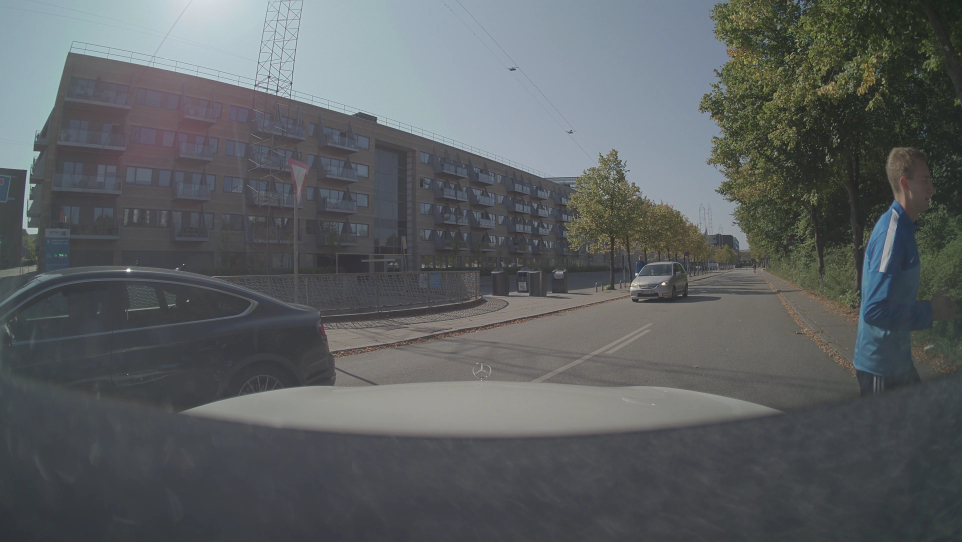}
    \includegraphics[width=0.19\linewidth]{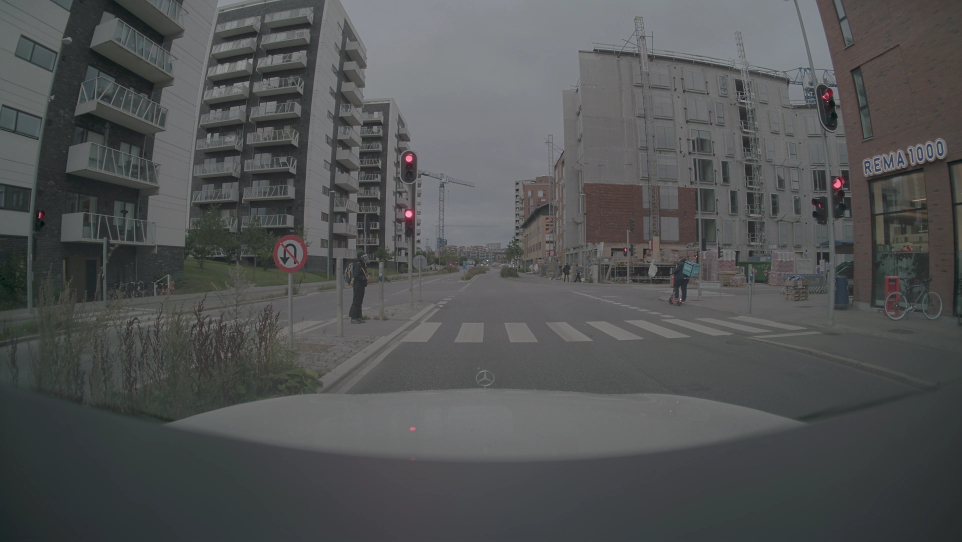}
    \includegraphics[width=0.19\linewidth]{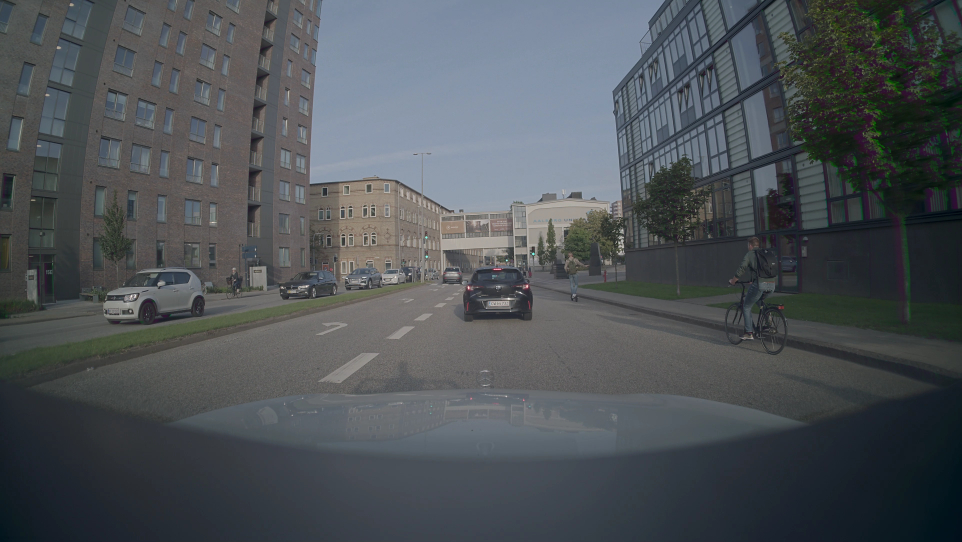}
    \includegraphics[width=0.19\linewidth]{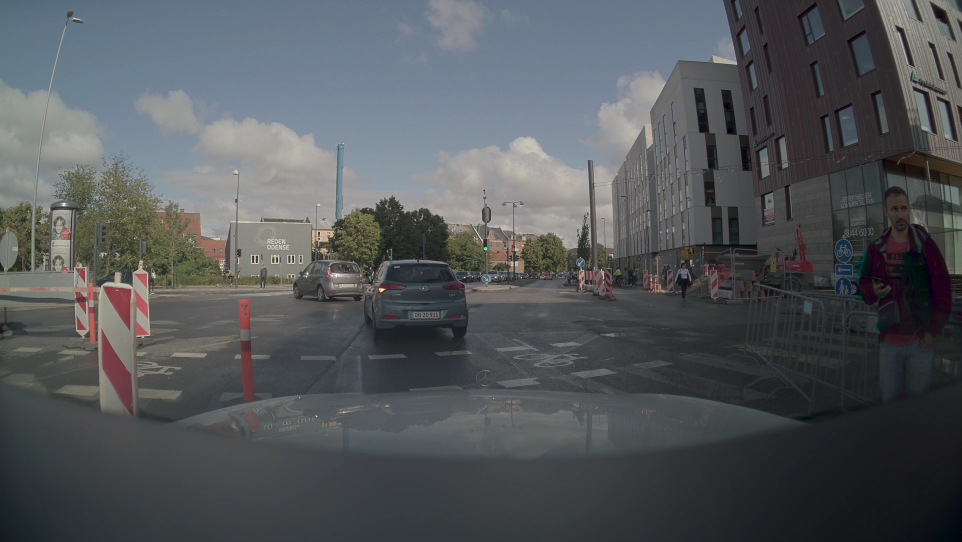}
    \includegraphics[width=0.19\linewidth]{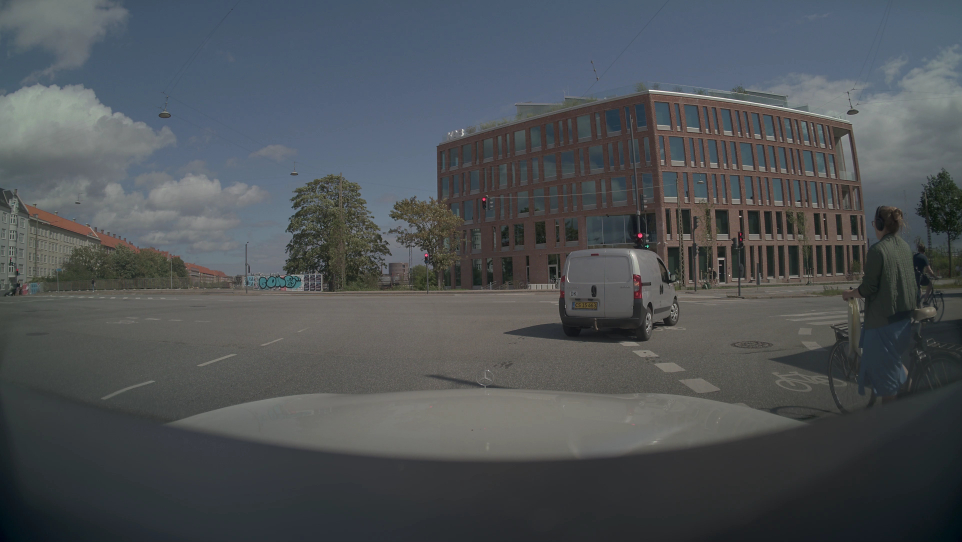} \\
    \includegraphics[width=0.19\linewidth]{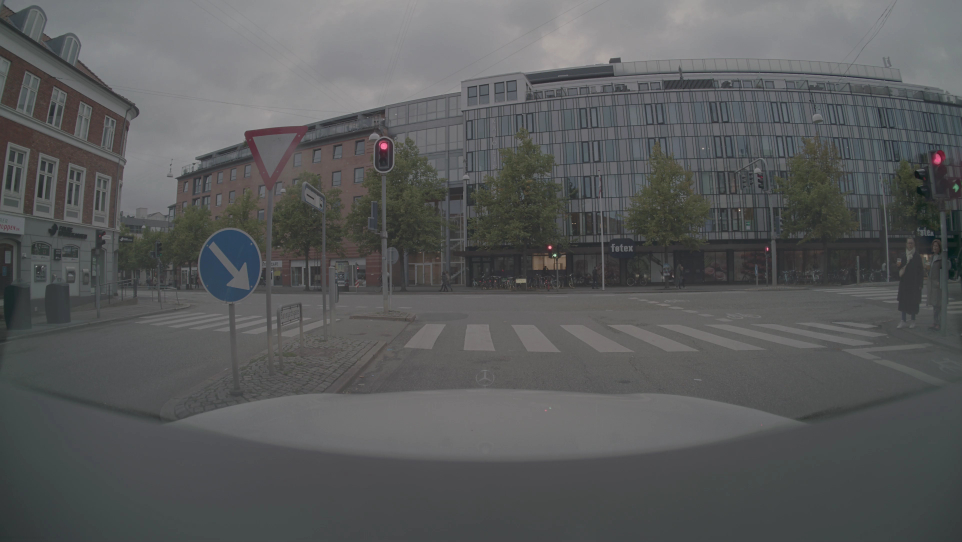}
    \includegraphics[width=0.19\linewidth]{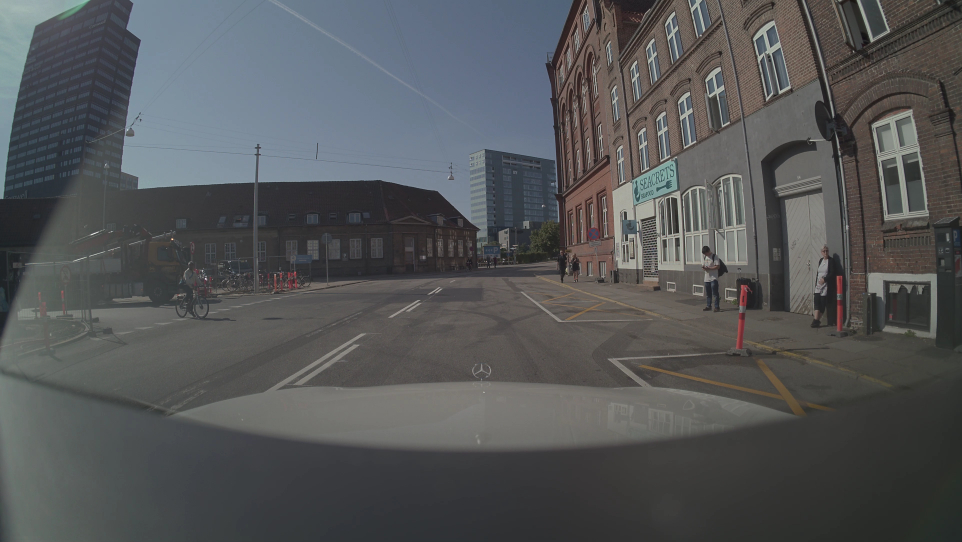}
    \includegraphics[width=0.19\linewidth]{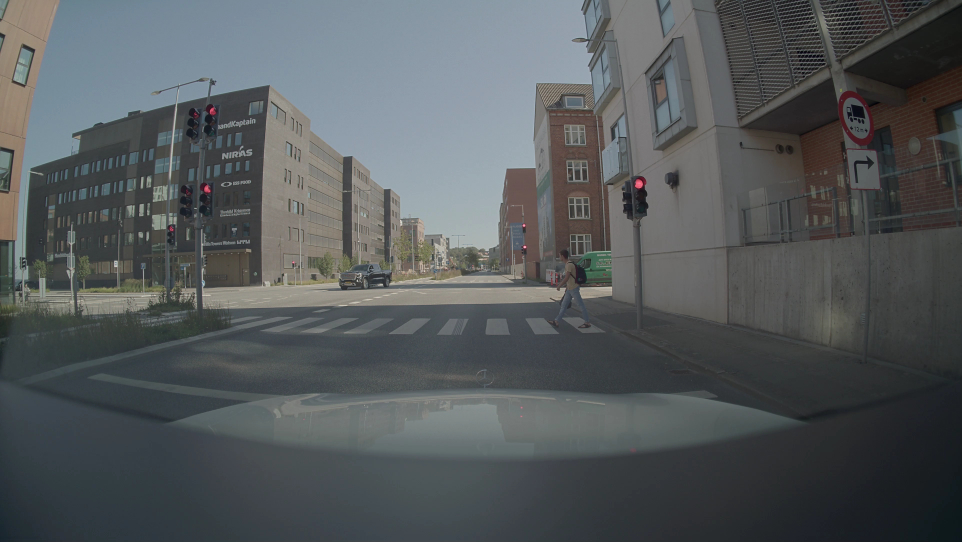}
    \includegraphics[width=0.19\linewidth]{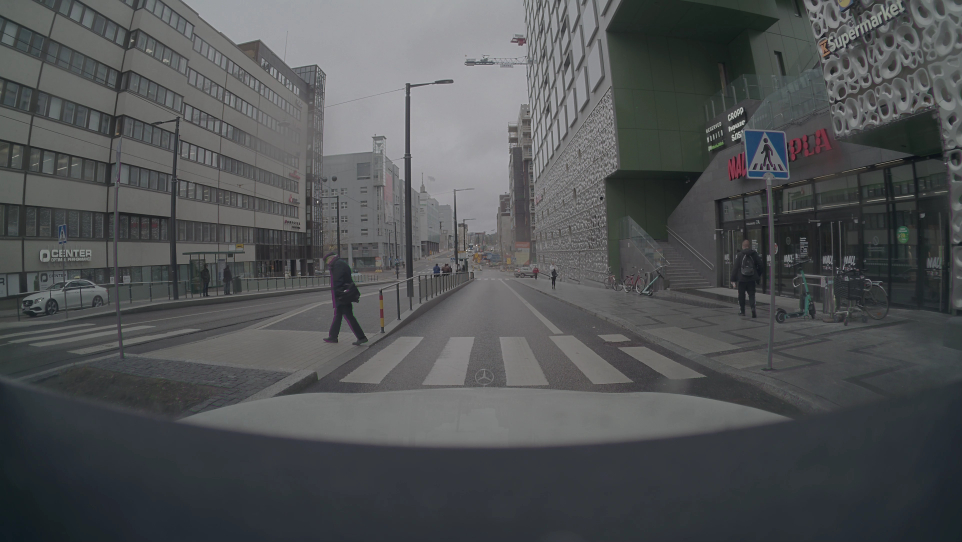}
    \includegraphics[width=0.19\linewidth]{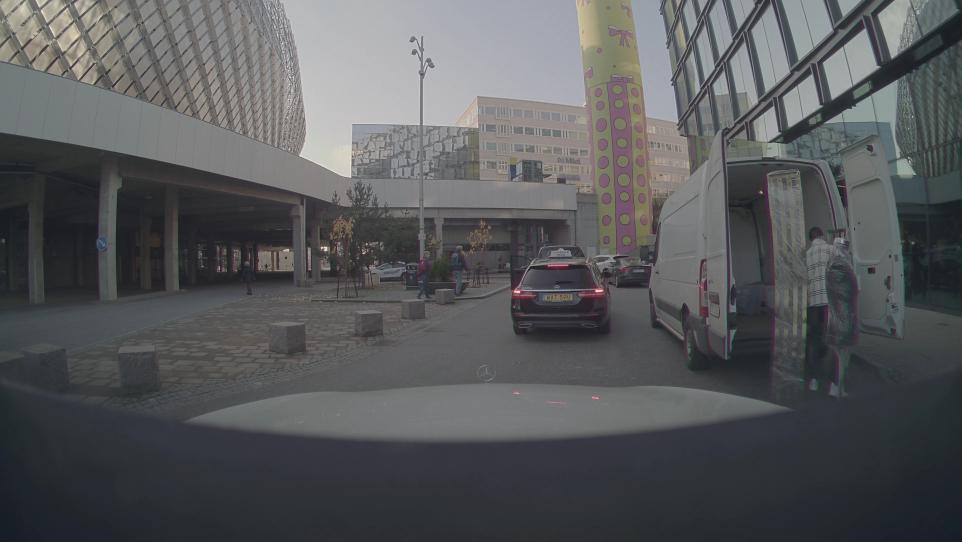}
    % \vspace{-0.4cm}
    \caption{Visualization of samples in one of our semantic clusters. The scenes are visually different but semantically similar (Pedestrians/cyclists near the ego car and likely to cross the street in front).}
    \label{fig:cluster-main-vis}
\end{figure*}

\paraq{Do Our Semantic Clusters Capture True Semantics}
First, we qualitatively inspect whether our semantical clusters contain scenes that are semantically and not merely visually similar. The cluster in~\cref{fig:cluster-main-vis} contains scenes with pedestrians or cyclists near the ego car that are likely to cross the street. Significantly, these scenes are visually diverse, taken at different times of day and locations.

Because we cannot visually inspect every cluster for semantic confirmation, we quantitatively analyze the composition of the dataset in ~\cref{fig:unique_sessions}. The original dataset is hand-curated to contain scenes from a large number of unique driving sessions. These different driving sessions tend to be geographically diverse. Thus, scenes within a session are visually more similar than scenes across sessions. Note that in our semantic clusters (LLaVA generated as shown in~\cref{sec:mllm}) contain a high number of unique sessions, which indicates that semantic clusters indeed capture semantically similar but visually diverse scenes across sessions. In contrast, clusters formed with CLIP visual embeddings have fewer unique sessions, indicating that visual clusters gather more visually similar scenes from a single session. This trend is stable and holds when we cluster the whole dataset, post data selection, and post data enrichment.

\begin{figure}[t]
    \centering
    \includegraphics[width=\linewidth]{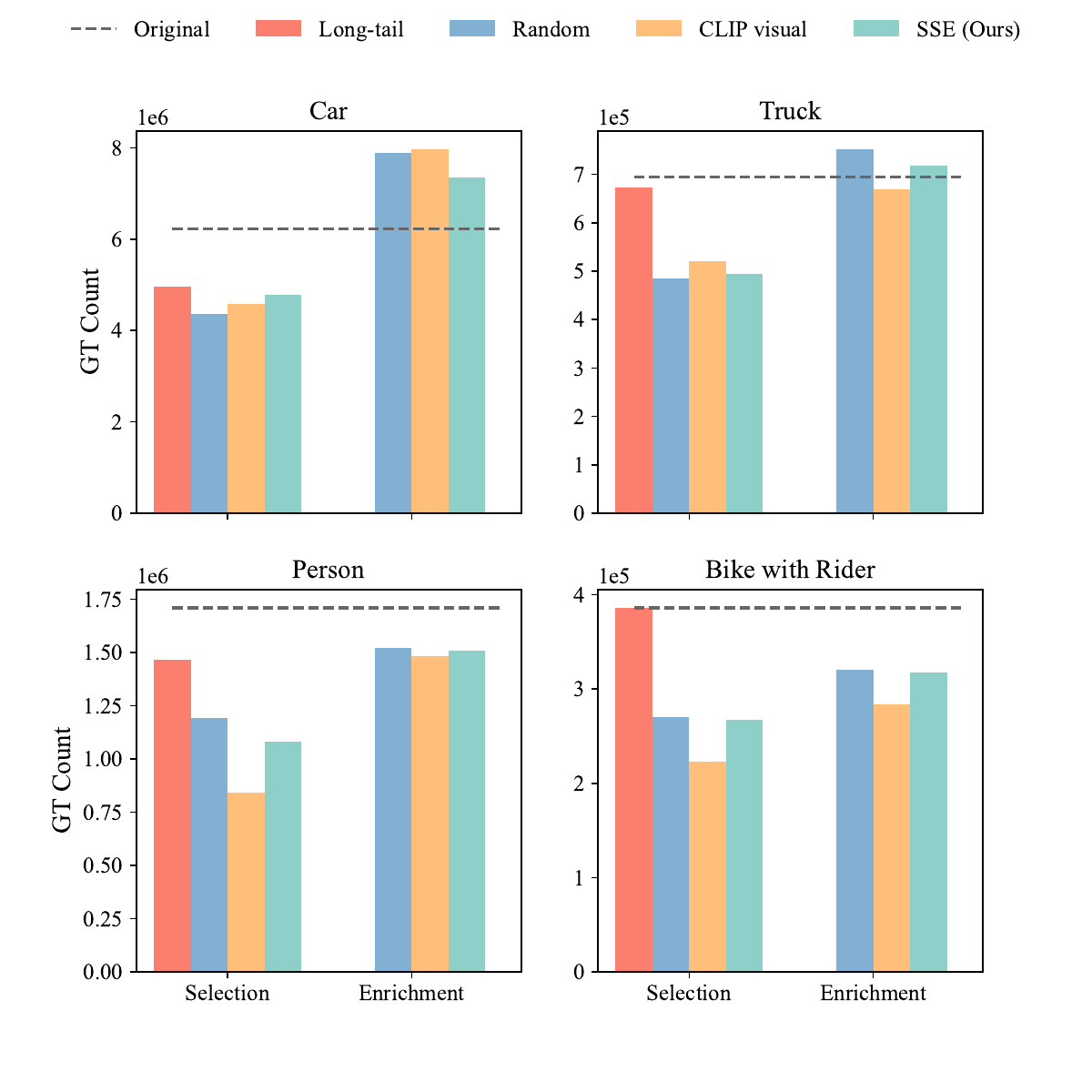}
    \caption{Number of unique objects after different selection and enrichment methods. SSE selects semantically diverse scenes that does not necessarily contain more rare objects.}
    \label{fig:gt_counts}
    \vspace{-.8cm}
\end{figure}

\begin{figure}
    \centering
    \includegraphics[width=0.95\linewidth]{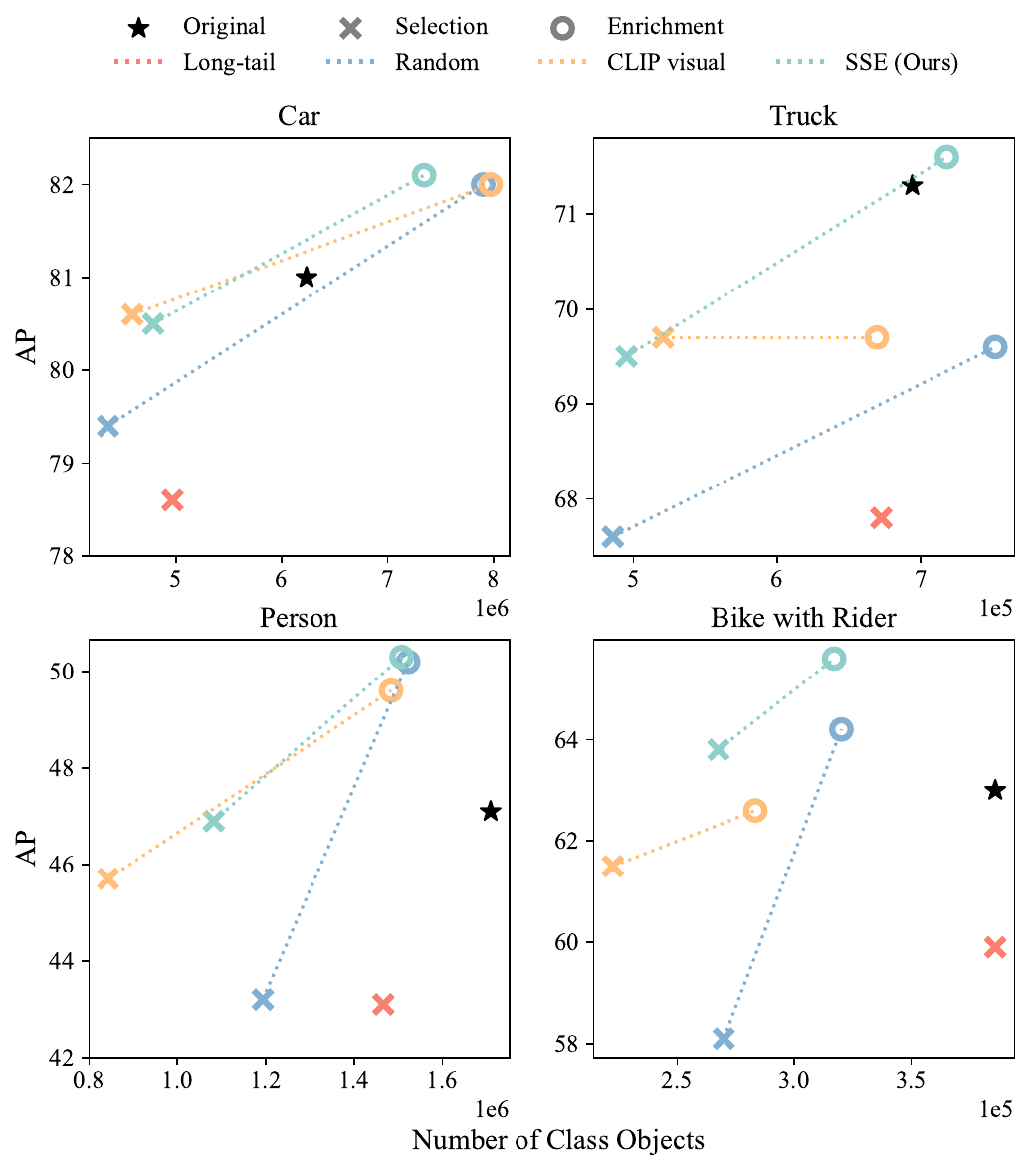}
    \caption{Per class detection accuracy as a function of object count in train data, across different methodologies. SSE semantically tunes the dataset and demonstrates that \textit{less but more high-quality objects} lead to better performance in rare categories.}
    \vspace{-0.5cm}
    \label{fig:gt_perf}
\end{figure}

\para{Semantically Diverse Data vs.\ Balanced Class Distribution Data}
We show how each methodology affects the number of unique ground truth objects. In our dataset, \texttt{car}, \texttt{truck}, \texttt{person}, and \texttt{bike with rider} cover the most to least represented objects. Interestingly, we observe that semantic selection and enrichment does not simply select and add images with rare objects, such as \texttt{person} and \texttt{bike with rider}, as seen in ~\cref{fig:gt_counts}. Importantly, we see in ~\cref{fig:gt_perf} that this imbalance in object distribution after selection does not severely affect the per-class AP. This makes it even more surprising that after enrichment the rarer categories with fewer objects outperform their original counterpart ($+3.2$ AP for \texttt{person}, $+2.6$ AP for \texttt{bike with rider}). Furthermore, SSE performs consistently better than all baselines across all categories after enrichment. Thus, SSE semantically tunes a dataset and successfully demonstrates that \textbf{fewer but more high-quality objects} lead to a better performance in rare categories. Summarily, we show that semantically diversified data is more important than resampled and balanced data.

% SSE does not select or add images with more rarer objects, but more semantically diverse scenes.

\subsection{Analyzing Effects of Parameters on Data Selection and Enrichment}
\begin{figure}[t]
    \centering
    \includegraphics[width=.8\linewidth]{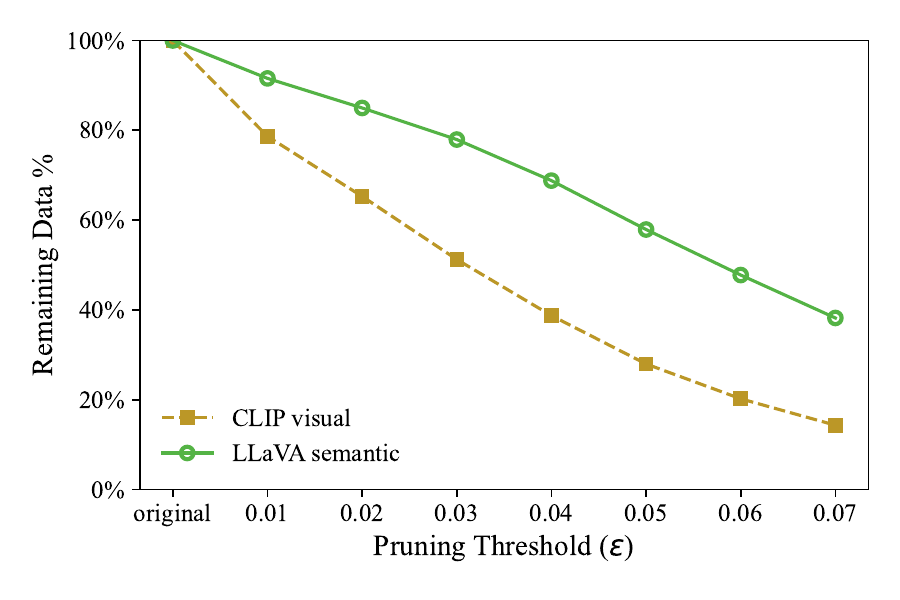}
    \vspace{-0.3cm}
    \caption{Fraction of data remaining vs pruning threshold $\epsilon$.}
    \label{fig:dedup} 
    % \vspace{-0.5cm}
\end{figure}

\para{Pruning Threshold $\epsilon$}
In data selection, we ensure visually diverse data within each semantic cluster through a pruning threshold $\epsilon$. Because the CLIP visual baseline uses the same visual embedding for clustering, we see that the pruning effects of $\epsilon$ are strongly pronounced, as shown in ~\cref{fig:dedup}. 
The effects of using different MLLMs are addressed later in~\cref{sec:mllm}. For all studies, we closely match the percentage of data across methodologies for fair comparison.

\begin{figure}[t]
    \centering
    \includegraphics[width=1\linewidth]{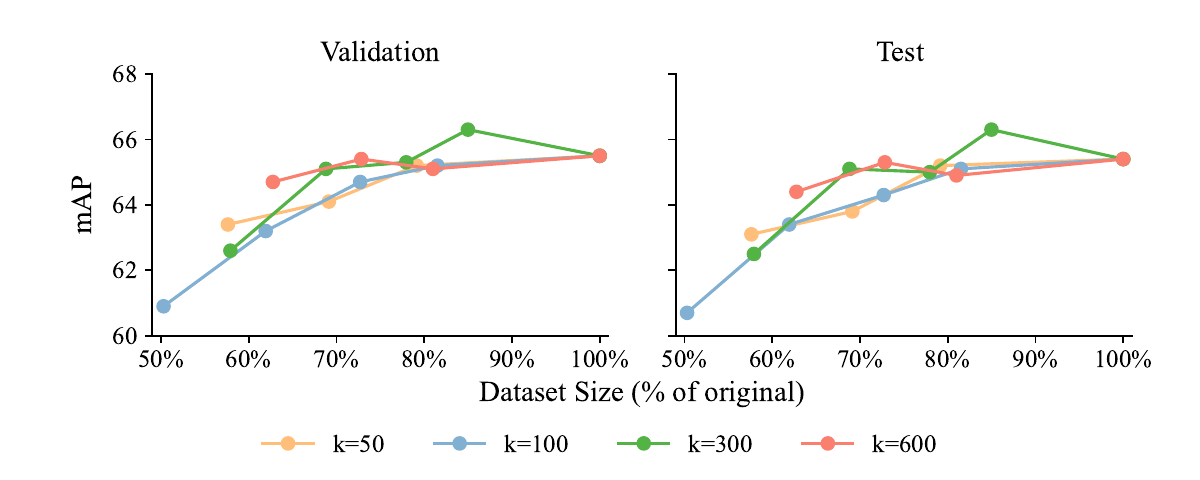}
    \caption{Study on the number of $k$ clusters.}
    \label{fig:k_ablation}
\end{figure}

\para{\textit{k} Selection}
Data selection and enrichment relies on clustering data embeddings with \textit{k}-means. We randomly hold out 20\% of the original dataset for validating \textit{k} clusters selection. We attempted to align the amount of retained data by setting similar pruning $\epsilon$.
As observed in ~\cref{fig:k_ablation}, we observe that semantic selection at $k=300$ is able to retain the fewest samples at $70\%$ with minimal performance change. Thus, we select $k=300$ as our default number of clusters for all experiments. Importantly, we observe that all data selection models' performances across the val and train set are approximately similar. This indicates that the effects of \textit{k} are stable across retention thresholds and data splits. 

\subsection{Studying the Effects of MLLMs}
\label{sec:mllm}
\begin{figure*}
    \centering
    \begin{subfigure}[t]{.27\textwidth}
        \centering
        \includegraphics[width=\linewidth]{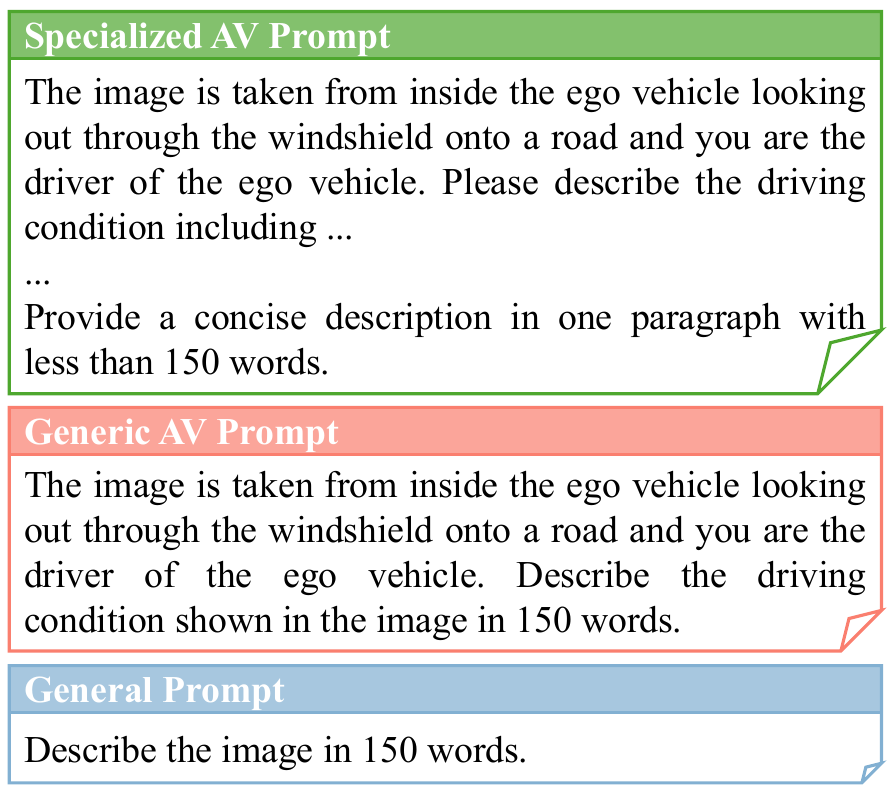}
        {(a) Three prompts } 
    \end{subfigure}
    \begin{subfigure}[t]{.65\textwidth}
        \centering
        \includegraphics[width=\linewidth]{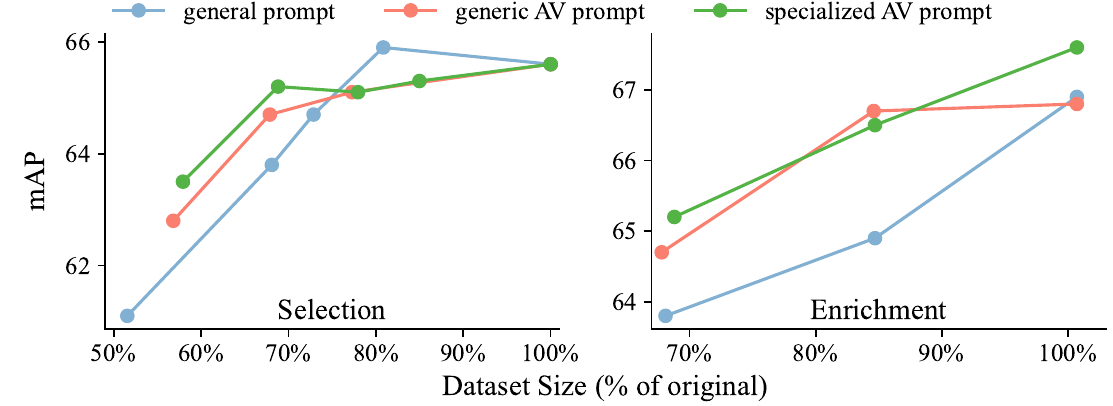}
        {(b) Dataset semantic selection and enrichment with different prompts.}
    \end{subfigure}
    % \vspace{-0.3cm}
    \caption{Study of prompt sensitivity. Specialized AV prompt provides guidance for selecting driving critical scenes.}
    \label{fig:prompt_sensitivity}
\end{figure*}

\para{Prompt Sensitivity}
The phenomenon of MLLM's sensitivity to prompts is well known and studied~\cite{mizrahi2024state,sclar2023quantifying,voronov2024mind,lu2021fantastically}. We attempt to find the best prompts suitable for semantic data selection and enrichment. The results for using different prompts for the MLLM are in~\cref{fig:prompt_sensitivity} for data selection and enrichment. In data selection, using an specialized AV prompt outperforms both the generic AV prompt and general prompt. The targeted questions within specialized AV prompt act as guildlines for selecting only the most critical scenes in the dataset. On the other hand, because we expand the dataset with data points that are more semantically diverse than our clusters, a more generic AV prompt is more befitting for a wider data search. Accordingly, we see that a generic AV prompt outperforms the rest. We report AV specialized prompt results as our main selection and enrichment results in~\cref{tab:main_selection,tab:main_enrich} because it best handles the data selection tradeoff in retention and performance. 

\begin{figure}[t]
    \centering

    \includegraphics[width=\linewidth]{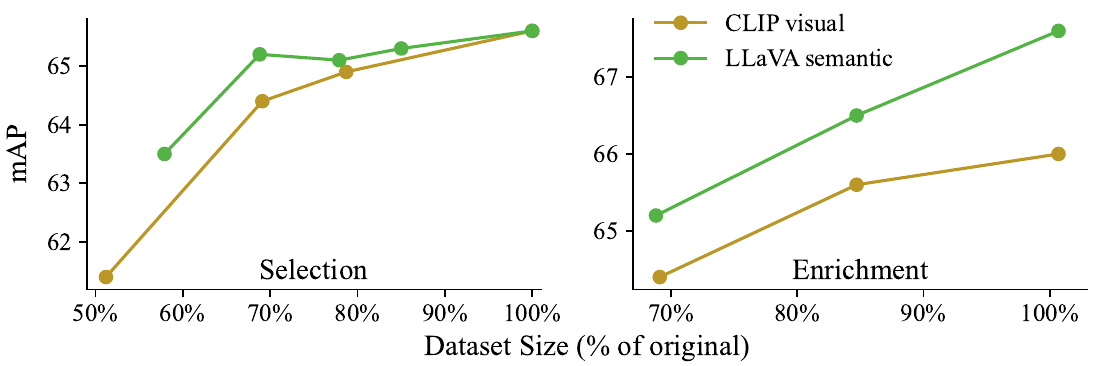}
    
    \caption{Study on the effects of different MLLMs.}
    \label{fig:mllms}
\end{figure}

\begin{figure*}[h]
\centering
\resizebox{\linewidth}{!}{
\begin{tabular}{c}
\includegraphics[width=0.97\linewidth,clip,trim= 0 0 0 0px]{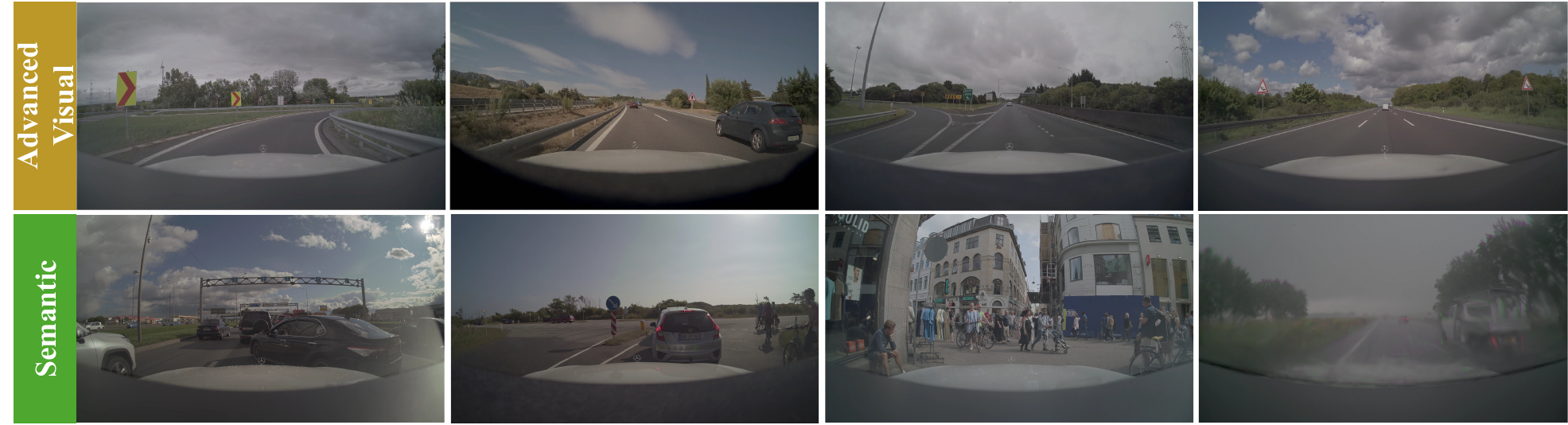} \\
    Query: Situations where need to slow down and drive carefully.
\end{tabular}
}
\vspace{-0.3cm}
\caption{Retrieval examples using advanced visual embeddings vs semantic embeddings. Semantic retrieval finds semantically similar yet critically visually diverse scenes. Visual retrieval returns visually repetitive scenes and cannot show high-level scene understanding.
}
\vspace{-0.2cm}
\label{fig:retrieval}
\end{figure*}

\para{MLLMs}
On our semantic data selection and enrichment, we compare the effects of different MLLMs for caption generation. Specifically we compare LLaVA with CLIP (vision language model). Significantly, we note that the performance trend in both semantic selection and enrichment is similar with both MLLMs, as seen in~\cref{fig:mllms}. We note that data retention for CLIP is notably lower because the default visual pruning process removes visually similar images based on CLIP embeddings.

\subsection{Semantic Retrieval}
Utilizing information rich captions as our semantic embeddings enables an additional capability: semantic retrieval. We set up a simple retrieval framework similar to standard retrieval tools powered by foundation models~\cite{radford2021CLIP}, where a similarity search is performed at a shared embedding space between a user prompt against a set of raw data points. We compare our retrievals against an advanced visual retrieval system, which uses CLIP vision encoder embeddings for patches of images~\cite{chang2023thinking}. As observed in~\cref{fig:retrieval}, semantic retrieval is able to retrieve semantically similar scenes that are visually diverse. For example, semantic retrieval is able to break down the concept of ``slow down and drive carefully'' and successfully retrieve various scene interpretations (starting from left to right images): vehicles cutting into lanes, bicyclist crossing, dense pedestrian traffic, and heavy rain. In comparison, CLIP retrievals are visually repetitive (mostly highway) and incapable of capturing high-level scene understanding beyond scenes with different traffic signs.

\section{Related Works}

\para{Data Pruning and Selection}
Selecting a good set of data points and pruning out bad data points are both considered even in the process of creating a dataset, prior to any model introduction. Public datasets, such as NuScenes, carefully balance the class and object count distributions and remove data points with common objects~\cite{caesar2020nuscenes}. However, this balancing requires considerable manual effort, is unscalable, and is entirely subjective. Recent large-scale datasets, such as LAION~\cite{schuhmann2022laion}, bypasses these issues by automatically removing visually similar images, as calculated through CLIP’s visual similarity score. Beyond dataset creation, data selection is commonly applied to existing datasets in order to reduce compute costs. Many approaches for classification datasets attempt to select the most diverse set of data points, either by maximizing visual space coverage~\cite{sener2017active} or maximizing gradient space coverage~\cite{ash2019deep}. 

In contrast, recent works have attempted to inject semantics into their data selection process, particularly through the use of foundation models like CLIP. Within the context of fairness, FAIRdedup~\cite{slyman2024fairdedup} uses CLIP to assign protected attributes for each image and attempts to balance the protected-attributes distribution. One analytical work shows that datasets pruned with optimal CLIP scores can improve model performance beyond normal datasets with the same distribution~\cite{sorscher2022beyond}. Another analytical work shows that under a computation budget, CLIP pruned dataset can lead to a higher quality dataset for better model performance~\cite{goyal2024scaling}. However, both works use brute force grid search of CLIP score thresholds to show these analytical insights and do not provide a solution for data selection. 

Finally, a series of studies (EL2N, GraNd, MoSo, Memorization, Forgetting score) assumes that hard samples are critical to datasets and aims to prune out easy samples~\cite{paul2021deep,tan2024data,feldman2020neural,toneva2018empirical}. They often require model ensembles or use downstream model itself to find hard samples. Thus, they are not computationally adaptable to industrial applications nor do they provide any explainability. In our work, we are able to semantically select data points from a large-scale dataset and provide interpretable selection reasons.

\para{Active Learning}
Similar to data pruning, active learning has been studied mostly in image classification. While some works target 2D object detection, they primarily work with small amounts of data to improve early stages of training~\cite{choi2021active,kao2019localization,yoo2019learning}. Within classification, active learning approaches range from uncertainty estimation over query by committee to training specialized models. Like data selection works, uncertainty estimation studies rely on the assumption that hard samples, for which the downstream model is uncertain in its predictions, are valuable for training~\cite{lakshminarayanan2017simple,liu2020simple,valdenegro1910deep,wen2020batchensemble,haussmann2020scalable}. Query by committee finds important data through majority vote of model ensembles, which requires even more models than uncertainty estimation~\cite{roy2018deep}. Finally, works that train specialized models to find their targeted data points only apply to settings where desired data types are provided~\cite{shu2023knowledge,bertoni2017data}. All these works are reliant on a model in the loop or additional models to find new useful data. This reliance raises compute costs, making them unscalable for industrial applications.

\section{Conclusion}
We introduced semantic selection and enrichment (SSE) which identifies the most semantically diverse and significant parts of a dataset and enhances it further by uncovering new data from a vast pool of unlabeled information. Importantly, SSE offers explainability by utilizing foundation models to generate semantics for each data point. Through semantic data selection, we show that we can select a semantical core group of data from an already carefully hand-curated, balanced, and human-labeled dataset. Our dataset post selection is only $70\%$ of the original dataset size but enables the downstream model to achieve similar performance to that from the original dataset. To expand this smaller semantic dataset back to $100\%$ of original dataset size, our enrichment process can successfully find and add more semantically meaningful data with increased downstream model performance. Crucially, we quantitatively show that the now semantically expanded dataset is able to improve rare class performance with even fewer rare but semantically important objects, demonstrating the strong advantage semantics imbue. Together, our semantic selection and enrichment contributes two main outcomes. First, we can reduce computational cost and achieve similar model performance. And second, a carefully handcrafted dataset can be semantically tuned without increasing the dataset size for better performance.

\section{Acknowledgements}
We would like to thank Jenny Schmalfuss for discussions and paper edits as well as Feiyang Kang for insightful comments and suggestions.

{\small
\bibliographystyle{ieee_fullname}
\bibliography{sample-base}

\begin{thebibliography}{10}\itemsep=-1pt

\bibitem{ash2019deep}
Jordan~T Ash, Chicheng Zhang, Akshay Krishnamurthy, John Langford, and Alekh Agarwal.
\newblock Deep batch active learning by diverse, uncertain gradient lower bounds.
\newblock {\em arXiv preprint arXiv:1906.03671}, 2019.

\bibitem{bertoni2017data}
Alessandro Bertoni and Tobias Larsson.
\newblock Data mining in product service systems design: Literature review and research questions.
\newblock {\em Procedia CIRP}, 64:306--311, 2017.

\bibitem{caesar2020nuscenes}
Holger Caesar, Varun Bankiti, Alex~H Lang, Sourabh Vora, Venice~Erin Liong, Qiang Xu, Anush Krishnan, Yu Pan, Giancarlo Baldan, and Oscar Beijbom.
\newblock nuscenes: A multimodal dataset for autonomous driving.
\newblock In {\em Proceedings of the IEEE/CVF conference on computer vision and pattern recognition}, pages 11621--11631, 2020.

\bibitem{chang2023thinking}
Nadine Chang, Francesco Ferroni, Michael~J Tarr, Martial Hebert, and Deva Ramanan.
\newblock Thinking like an annotator: Generation of dataset labeling instructions.
\newblock {\em arXiv preprint arXiv:2306.14035}, 2023.

\bibitem{choi2021active}
Jiwoong Choi, Ismail Elezi, Hyuk-Jae Lee, Clement Farabet, and Jose~M Alvarez.
\newblock Active learning for deep object detection via probabilistic modeling.
\newblock In {\em ICCV}, pages 10264--10273, 2021.

\bibitem{feldman2020neural}
Vitaly Feldman and Chiyuan Zhang.
\newblock What neural networks memorize and why: Discovering the long tail via influence estimation.
\newblock {\em NeurIPS}.

\bibitem{goyal2024scaling}
Sachin Goyal, Pratyush Maini, Zachary~C Lipton, Aditi Raghunathan, and J~Zico Kolter.
\newblock Scaling laws for data filtering--data curation cannot be compute agnostic.
\newblock In {\em CVPR}, pages 22702--22711, 2024.

\bibitem{gupta2019lvis}
Agrim Gupta, Piotr Dollar, and Ross Girshick.
\newblock Lvis: A dataset for large vocabulary instance segmentation.
\newblock In {\em CVPR}, pages 5356--5364, 2019.

\bibitem{haussmann2020scalable}
Elmar Haussmann, Michele Fenzi, Kashyap Chitta, Jan Ivanecky, Hanson Xu, Donna Roy, Akshita Mittel, Nicolas Koumchatzky, Clement Farabet, and Jose~M Alvarez.
\newblock Scalable active learning for object detection.
\newblock In {\em 2020 IEEE intelligent vehicles symposium}, pages 1430--1435. IEEE, 2020.

\bibitem{kao2019localization}
Chieh-Chi Kao, Teng-Yok Lee, Pradeep Sen, and Ming-Yu Liu.
\newblock Localization-aware active learning for object detection.
\newblock In {\em ACCV}, pages 506--522. Springer, 2019.

\bibitem{lakshminarayanan2017simple}
Balaji Lakshminarayanan, Alexander Pritzel, and Charles Blundell.
\newblock Simple and scalable predictive uncertainty estimation using deep ensembles.
\newblock {\em NeurIPS}, 30, 2017.

\bibitem{liu2023improvedllava}
Haotian Liu, Chunyuan Li, Yuheng Li, and Yong~Jae Lee.
\newblock Improved baselines with visual instruction tuning, 2023.

\bibitem{liu2020simple}
Jeremiah Liu, Zi Lin, Shreyas Padhy, Dustin Tran, Tania Bedrax~Weiss, and Balaji Lakshminarayanan.
\newblock Simple and principled uncertainty estimation with deterministic deep learning via distance awareness.
\newblock {\em NeurIPS}, 33:7498--7512, 2020.

\bibitem{lu2021fantastically}
Yao Lu, Max Bartolo, Alastair Moore, Sebastian Riedel, and Pontus Stenetorp.
\newblock Fantastically ordered prompts and where to find them: Overcoming few-shot prompt order sensitivity.
\newblock {\em arXiv preprint arXiv:2104.08786}, 2021.

\bibitem{mizrahi2024state}
Moran Mizrahi, Guy Kaplan, Dan Malkin, Rotem Dror, Dafna Shahaf, and Gabriel Stanovsky.
\newblock State of what art? a call for multi-prompt llm evaluation.
\newblock {\em Transactions of the Association for Computational Linguistics}, 12:933--949, 2024.

\bibitem{paul2021deep}
Mansheej Paul, Surya Ganguli, and Gintare~Karolina Dziugaite.
\newblock Deep learning on a data diet: Finding important examples early in training.
\newblock {\em NeurIPS}, 34:20596--20607, 2021.

\bibitem{pham2023nvautonet}
Trung Pham, Mehran Maghoumi, Wanli Jiang, Bala Siva~Sashank Jujjavarapu, Mehdi Sajjadi, Xin Liu, Hsuan-Chu Lin, Bor-Jeng Chen, Giang Truong, Chao Fang, et~al.
\newblock Nvautonet: Fast and accurate 360 3d visual perception for self driving.
\newblock {\em arXiv preprint arXiv:2303.12976}, 2023.

\bibitem{radford2021CLIP}
Alec Radford, Jong~Wook Kim, Chris Hallacy, Aditya Ramesh, Gabriel Goh, Sandhini Agarwal, Girish Sastry, Amanda Askell, Pamela Mishkin, Jack Clark, et~al.
\newblock Learning transferable visual models from natural language supervision.
\newblock In {\em International conference on machine learning}, pages 8748--8763. PMLR, 2021.

\bibitem{roy2018deep}
Soumya Roy, Asim Unmesh, and Vinay~P Namboodiri.
\newblock Deep active learning for object detection.
\newblock In {\em BMVC}, volume 362, page~91, 2018.

\bibitem{schuhmann2022laion}
Christoph Schuhmann, Romain Beaumont, Richard Vencu, Cade Gordon, Ross Wightman, Mehdi Cherti, Theo Coombes, Aarush Katta, Clayton Mullis, Mitchell Wortsman, et~al.
\newblock Laion-5b: An open large-scale dataset for training next generation image-text models.
\newblock {\em Advances in Neural Information Processing Systems}, 35:25278--25294, 2022.

\bibitem{sclar2023quantifying}
Melanie Sclar, Yejin Choi, Yulia Tsvetkov, and Alane Suhr.
\newblock Quantifying language models' sensitivity to spurious features in prompt design or: How i learned to start worrying about prompt formatting.
\newblock {\em arXiv preprint arXiv:2310.11324}, 2023.

\bibitem{sener2017active}
Ozan Sener and Silvio Savarese.
\newblock Active learning for convolutional neural networks: A core-set approach.
\newblock {\em arXiv preprint arXiv:1708.00489}, 2017.

\bibitem{shu2023knowledge}
Xiaoling Shu and Yiwan Ye.
\newblock Knowledge discovery: Methods from data mining and machine learning.
\newblock {\em Social Science Research}, 110:102817, 2023.

\bibitem{slyman2024fairdedup}
Eric Slyman, Stefan Lee, Scott Cohen, and Kushal Kafle.
\newblock Fairdedup: Detecting and mitigating vision-language fairness disparities in semantic dataset deduplication.
\newblock In {\em CVPR}, pages 13905--13916, 2024.

\bibitem{smith2019super}
Leslie~N Smith and Nicholay Topin.
\newblock Super-convergence: Very fast training of neural networks using large learning rates.
\newblock In {\em Artificial intelligence and machine learning for multi-domain operations applications}, volume 11006, pages 369--386. SPIE, 2019.

\bibitem{song2020mpnet}
Kaitao Song, Xu Tan, Tao Qin, Jianfeng Lu, and Tie-Yan Liu.
\newblock Mpnet: Masked and permuted pre-training for language understanding.
\newblock {\em Advances in neural information processing systems}, 33:16857--16867, 2020.

\bibitem{sorscher2022beyond}
Ben Sorscher, Robert Geirhos, Shashank Shekhar, Surya Ganguli, and Ari Morcos.
\newblock Beyond neural scaling laws: beating power law scaling via data pruning.
\newblock {\em NeurIPS}, 35:19523--19536, 2022.

\bibitem{tan2024data}
Haoru Tan, Sitong Wu, Fei Du, Yukang Chen, Zhibin Wang, Fan Wang, and Xiaojuan Qi.
\newblock Data pruning via moving-one-sample-out.
\newblock {\em NeurIPS}, 36, 2024.

\bibitem{toneva2018empirical}
Mariya Toneva, Alessandro Sordoni, Remi Tachet~des Combes, Adam Trischler, Yoshua Bengio, and Geoffrey~J Gordon.
\newblock An empirical study of example forgetting during deep neural network learning.
\newblock {\em arXiv preprint arXiv:1812.05159}, 2018.

\bibitem{valdenegro1910deep}
Matias Valdenegro-Toro.
\newblock Deep sub-ensembles for fast uncertainty estimation in image classification. arxiv preprint arxiv: 191008168.(2019) doi: 10.48550.
\newblock {\em arXiv}, 1910.

\bibitem{voronov2024mind}
Anton Voronov, Lena Wolf, and Max Ryabinin.
\newblock Mind your format: Towards consistent evaluation of in-context learning improvements.
\newblock {\em arXiv preprint arXiv:2401.06766}, 2024.

\bibitem{wen2020batchensemble}
Yeming Wen, Dustin Tran, and Jimmy Ba.
\newblock Batchensemble: an alternative approach to efficient ensemble and lifelong learning.
\newblock {\em arXiv preprint arXiv:2002.06715}, 2020.

\bibitem{yoo2019learning}
Donggeun Yoo and In~So Kweon.
\newblock Learning loss for active learning.
\newblock In {\em CVPR}, pages 93--102, 2019.

\end{thebibliography}
}

\clearpage
\appendix

\begin{figure*}[h!]
    \centering
    \includegraphics[width=0.19\linewidth]{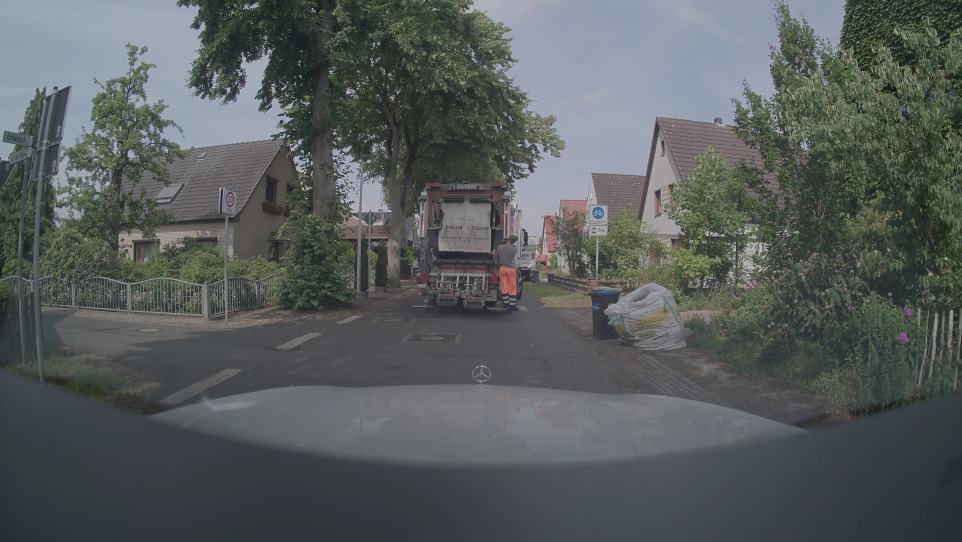}
    \includegraphics[width=0.19\linewidth]{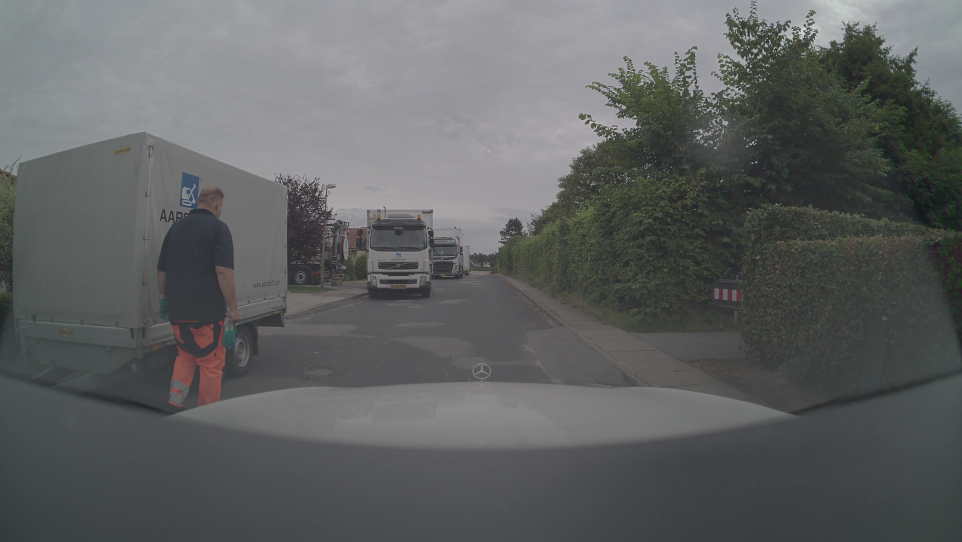}
    \includegraphics[width=0.19\linewidth]{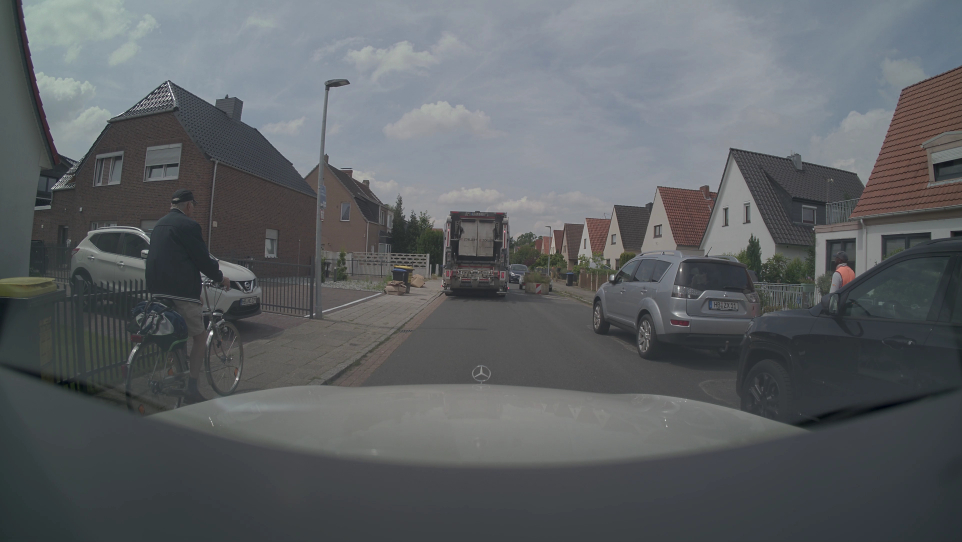}
    \includegraphics[width=0.19\linewidth]{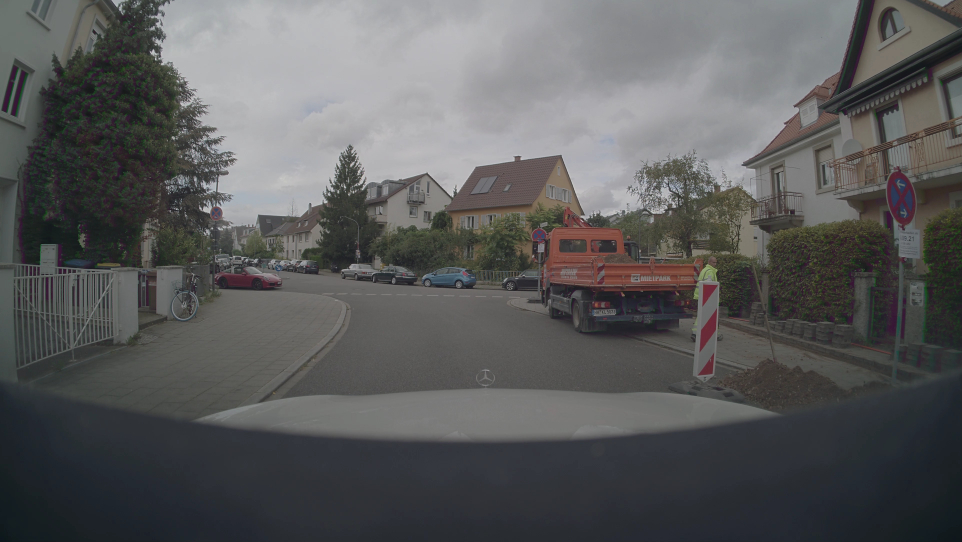}
    \includegraphics[width=0.19\linewidth]{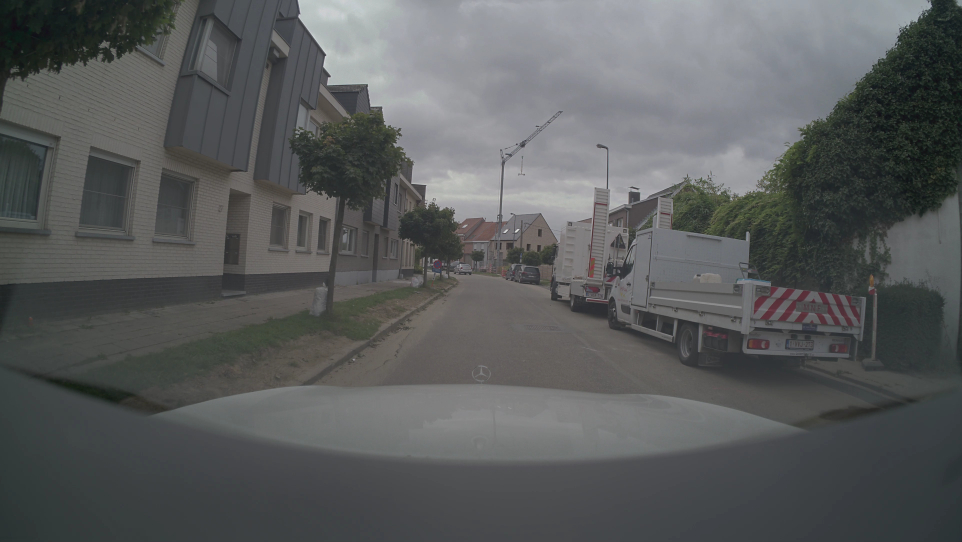} \\
    \includegraphics[width=0.19\linewidth]{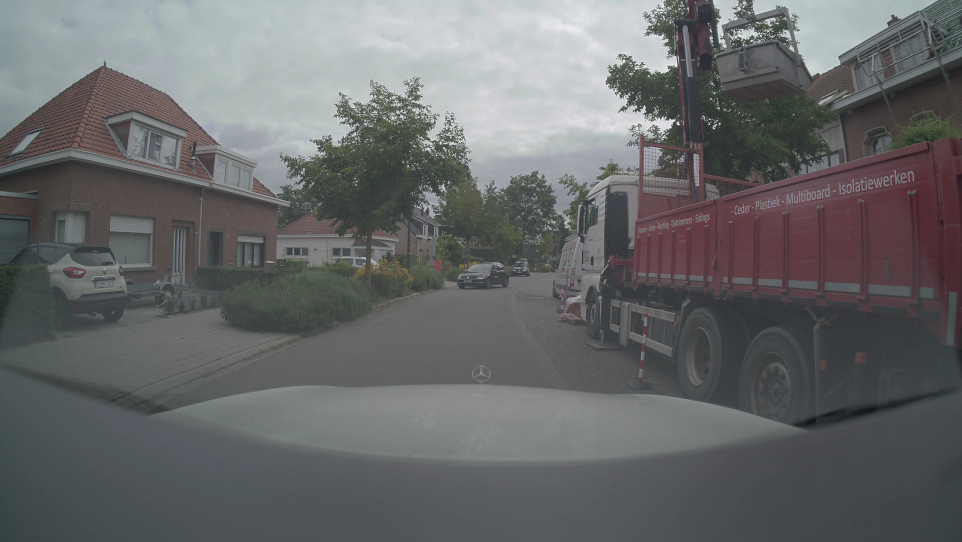}
    \includegraphics[width=0.19\linewidth]{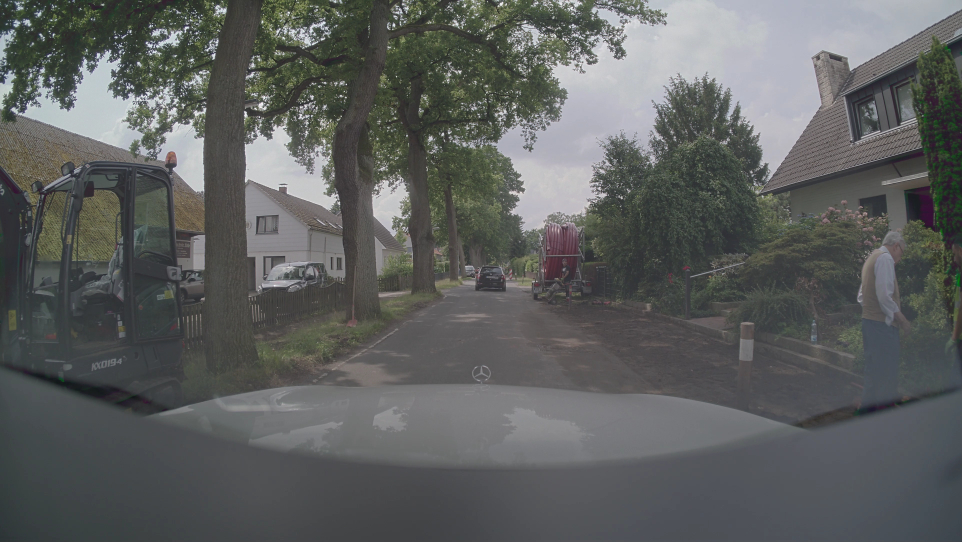}
    \includegraphics[width=0.19\linewidth]{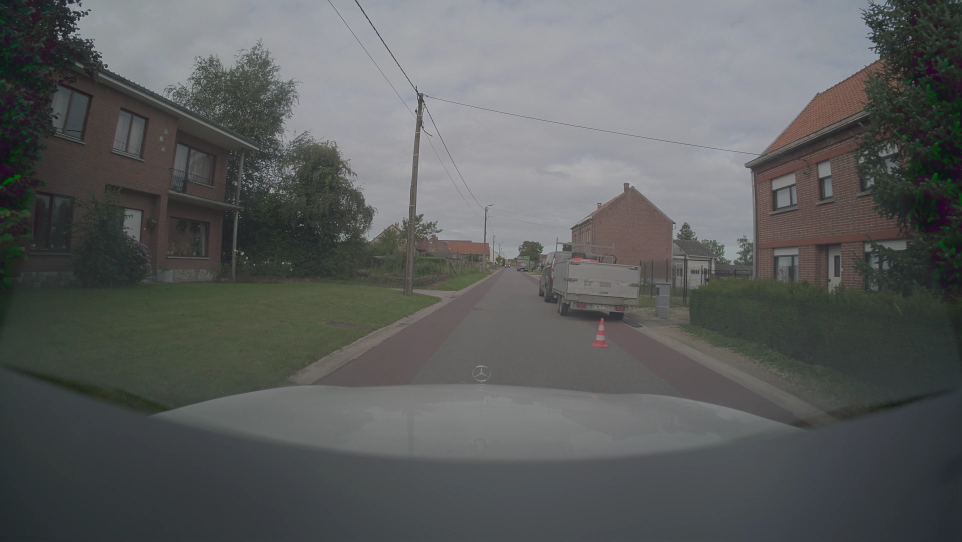}
    \includegraphics[width=0.19\linewidth]{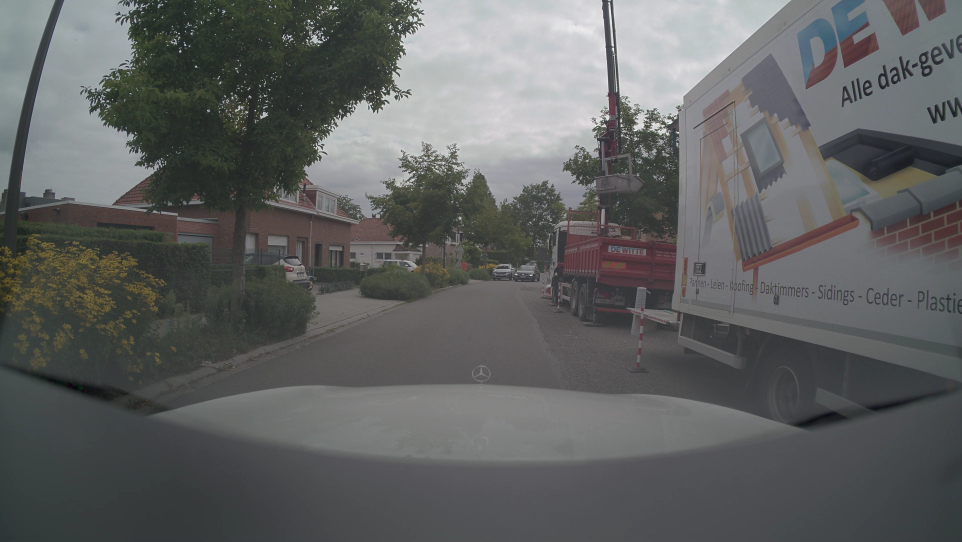}
    \includegraphics[width=0.19\linewidth]{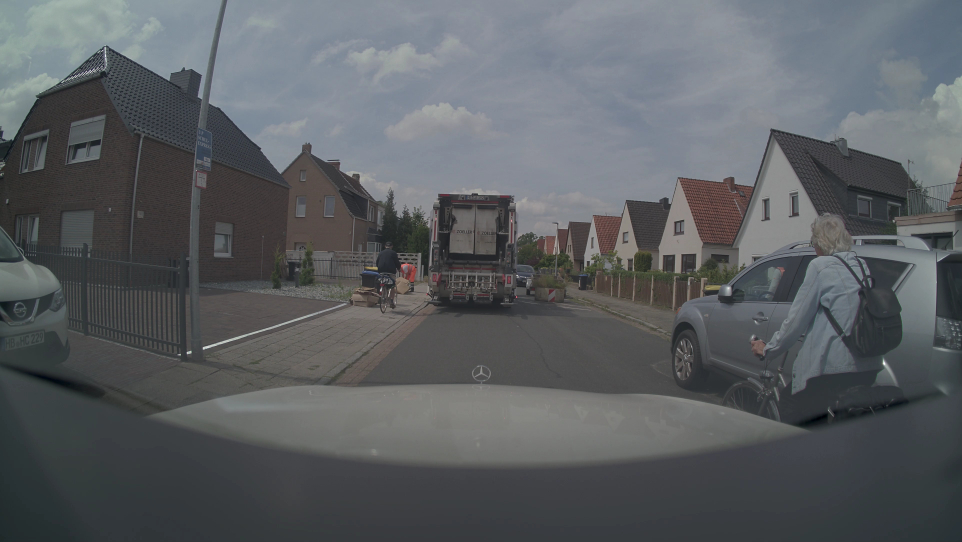}
    \caption{Visualization of samples in a cluster. The scenes are visually different but semantically similar. (Construction car/public service car, with potential worker/pedestrian)}
    \label{fig:cluster-0-vis}
\end{figure*}

\begin{figure*}[h!]
\centering
\resizebox{\linewidth}{!}{
\begin{tabular}{c}
    \includegraphics[width=0.98\linewidth,clip,trim= 0 0 0 0px]{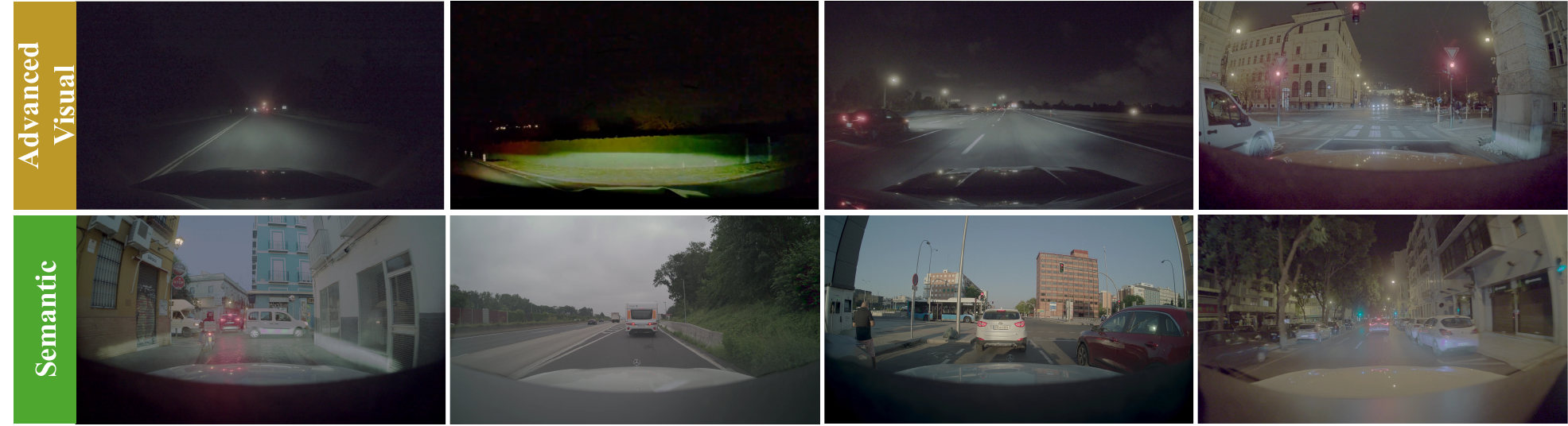} \\
    Query: The car in front stops with breaking lights on.
\end{tabular}
}
\label{fig:retrieval-appendix}
\caption{Additional retrieval examples using advanced visual embeddings and semantic embeddings.
}
\label{fig:retrieval_appendix}
\end{figure*}

\section{Additional Cluster Visualizations}
We provide additional cluster visualization in ~\cref{fig:cluster-0-vis}. The cluster contains scenes where there are construction car or public service car near the ego car, likely with workers or pedestrians. 
\label{appendix}

\section{Semantic Retrieval Visualizations}
We provide additional examples of retrieval in ~\cref{fig:retrieval_appendix}. Semantic retrieval is able to understand the more complex semantics "car stops" and "breaking lights on", while CLIP retrieval focus on "lights on" and returns mostly night images with lights on.

\section{Detailed Per-Class AP}
We provide detailed numbers for mAP and per-class AP for our main results in ~\cref{tab:detailed-ap}.
\begin{table*}[!t]
    \centering
    \caption{Per-class AP of 3D object detection with different data strategies.}
    \begin{tabular}{c|ccccccc}
    \toprule
     & Dataset Size & \multirow{2}{*}{Method} & \multirow{2}{*}{mAP} & \multicolumn{4}{c}{Per-Class AP} \\
     \cmidrule{5-8}
     & ($\%$ of original) & & & Car & Truck & Person & Bike with Rider \\
     \cmidrule{2-8}
     & $100\%$ & Original & $65.6$ & $81.0$ & $71.3$ & $47.1$ & $63.0$ \\
     \midrule
    \multirow{4}{*}{\rotatebox[origin=c]{90}{Selection}} & \multirow{4}{*}{$70\%$} & Random & $62.1$ & $79.4$ & $67.6$ & $43.2$ & $58.1$ \\
    & & Long-tail~\cite{gupta2019lvis} &  $62.3$ & $78.6$ & $67.8$ & $43.1$ & $59.9$\\
     & & CLIP visual~\cite{radford2021CLIP} & $64.4$ & $\mathbf{80.6}$ & $\mathbf{69.7}$ & $45.7$ & $61.5$ \\
     & & \textbf{SSD (Ours)} & $\mathbf{65.2}$ & $80.5$ & $69.5$ & $\mathbf{46.9}$ & $\mathbf{63.8}$ \\
     \midrule
     \multirow{3}{*}{\rotatebox[origin=c]{90}{Enrichment}} & \multirow{3}{*}{$100\%$} & Random & $66.5$ & $82.0$ & $69.6$ & $50.2$ & $64.2$ \\
     & & Long-tail~\cite{gupta2019lvis} & - & - & - & - & - \\
     & & CLIP visual~\cite{radford2021CLIP} & $66.0$ & $82.0$ & $69.7$ & $49.6$ & $62.6$ \\
     & & \textbf{SSD (Ours)} & $\mathbf{67.6}$ & $\mathbf{82.1}$ & $\mathbf{71.6}$ & $\mathbf{50.3}$ & $\mathbf{65.6}$ \\
     \bottomrule
    \end{tabular}
    \label{tab:detailed-ap}
\end{table*}

\end{document}